\documentclass[letterpaper]{article} 
\usepackage[preprint]{aaai2027}
\usepackage[hyphens]{url}  
\usepackage{graphicx} 
\usepackage{natbib}  
\usepackage{caption} 
\usepackage{amsmath}
\usepackage{amssymb}
\usepackage{multirow}
\usepackage{booktabs}
\usepackage{array}
\usepackage{colortbl}
\usepackage{xcolor}
\usepackage{tikz}
\usepackage{pifont}
\usepackage{bm}
\usepackage{subcaption}
\usepackage{afterpage}
\usepackage{float}

\begin{document}

\raggedbottom

\title{Where Detectors Fail: Closing the Tail-Domain Gap with Expert-Guided Mutual Distillation}

\author{
Xuan Feng\textsuperscript{1,3},
Guihong Liu\textsuperscript{2},
Tianlong Gu\textsuperscript{1}\corresponding,
Shuai Zhao\textsuperscript{3},\\
Xuemin Wang\textsuperscript{2},
Chenzhong Bin\textsuperscript{2},
Yang Liu\textsuperscript{4},
Bo An\textsuperscript{3}
}
\affiliations{
\textsuperscript{1}Jinan University \\
\textsuperscript{2}Guilin University of Electronic Technology \\
\textsuperscript{3}Nanyang Technological University \\
\textsuperscript{4}The Hong Kong Polytechnic University \\
\texttt{fenffef@gmail.com, gutianlong@jnu.edu.cn}
}

\maketitle

\begin{abstract}
Multimodal fake news detectors often generalize poorly across domains because they learn to trust unreliable evidence: domain-specific shortcuts amplified by imbalanced data and semantically inconsistent text--image pairs that make cross-modal evidence unreliable. We propose Expert-Guided Mutual Distillation (\textsc{EGMD}), which learns what evidence to trust across the prediction pipeline. At the input level, input-level calibration encodes pair-level coherence as a shared gain before fusion. At the representation level, an expert-guided teacher aligns domain statistics and encourages domain-specific patterns to concentrate in specialized experts. At the decision level, prototype-anchored domain-specific students use mutual learning and dual-channel distillation to inherit the teacher's feature geometry and calibrated predictions while discouraging local domain priors. We further construct \textsc{Weibo\_Balanced}, a domain-balanced benchmark that isolates the effect of imbalance on generalization. Across four datasets in two languages, \textsc{EGMD} achieves state-of-the-art accuracy while reducing domain bias by up to 57.3\%.
\end{abstract}

\newcommand{\scorebar}[2]{%
  \begin{tikzpicture}[baseline=-0.5ex]
    \fill[gray!20] (0,0) rectangle (4,0.4);
    \fill[#1] (0,0) rectangle (#2*4,0.4);
    \draw[dashed, thick] (2,-0.1) -- (2,0.5);
    \node[font=\tiny, anchor=east] at (-0.1,0.2) {0:Real};
    \node[font=\tiny, anchor=south] at (2,0.5) {0.5};
    \node[font=\tiny, anchor=west] at (4.1,0.2) {1:Fake};
    \draw[thick] (#2*4,0) -- (#2*4,0.4);
    \node[font=\tiny, anchor=south] at (#2*4,0.4) {#2};
  \end{tikzpicture}%
}

\newcommand{\semanticmisalignmentfigure}{%
\begin{figure}[t]
    \centering
    \setlength{\tabcolsep}{1pt}
    \scriptsize
    \renewcommand{\arraystretch}{1.15}

    \textbf{Health News:} COVID-19 vaccination safe for pregnant people...

    \vspace{2pt}
    \begin{tabular}{>{\centering\arraybackslash}m{0.49\linewidth}
                    >{\centering\arraybackslash}m{0.49\linewidth}}

    \textbf{(i) Matched: \textcolor{green!50!black}{Health Image}} &
    \textbf{(ii) Mismatched: \textcolor{red!70!black}{Politics Image}} \\

    \includegraphics[width=1.0\linewidth]{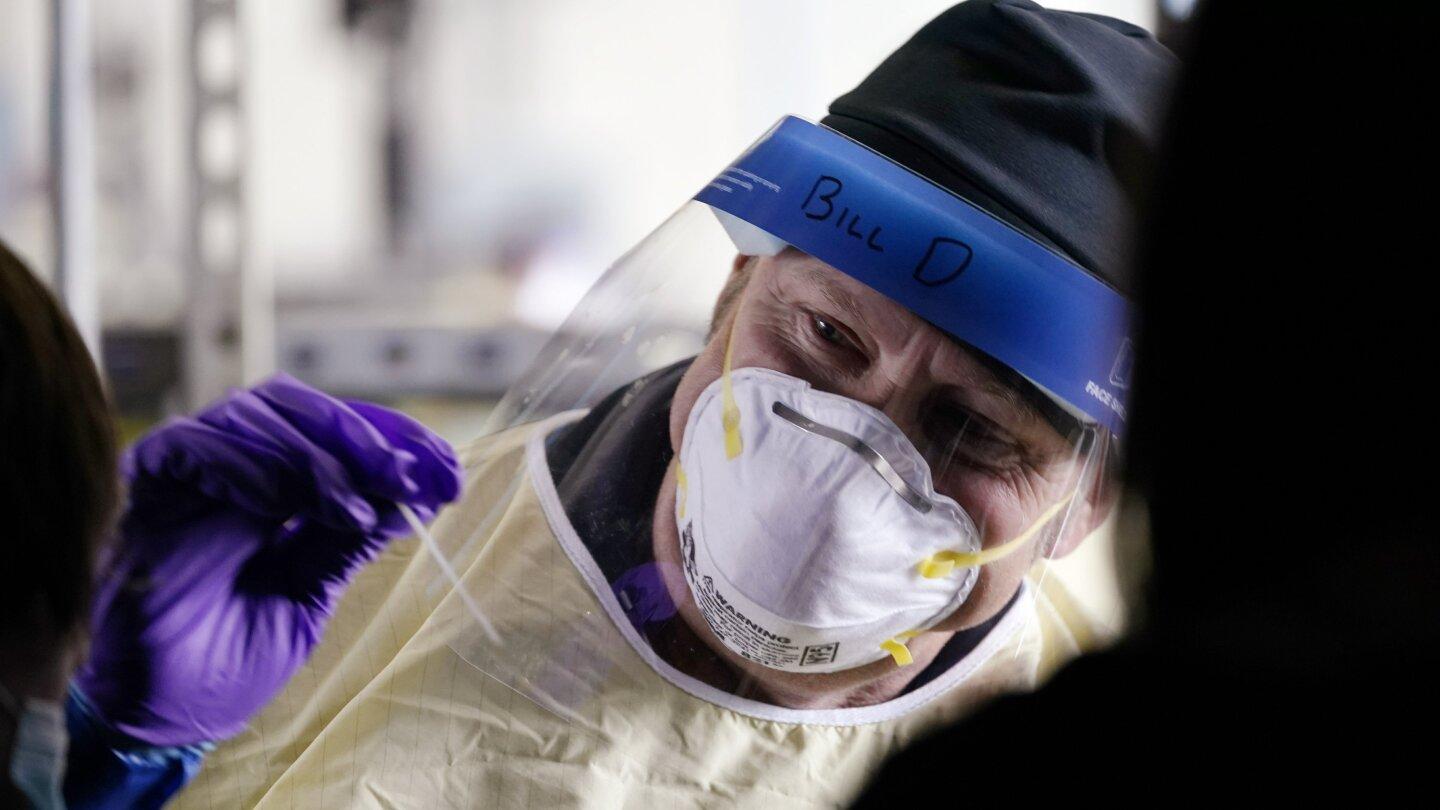} &
    \includegraphics[width=1.0\linewidth]{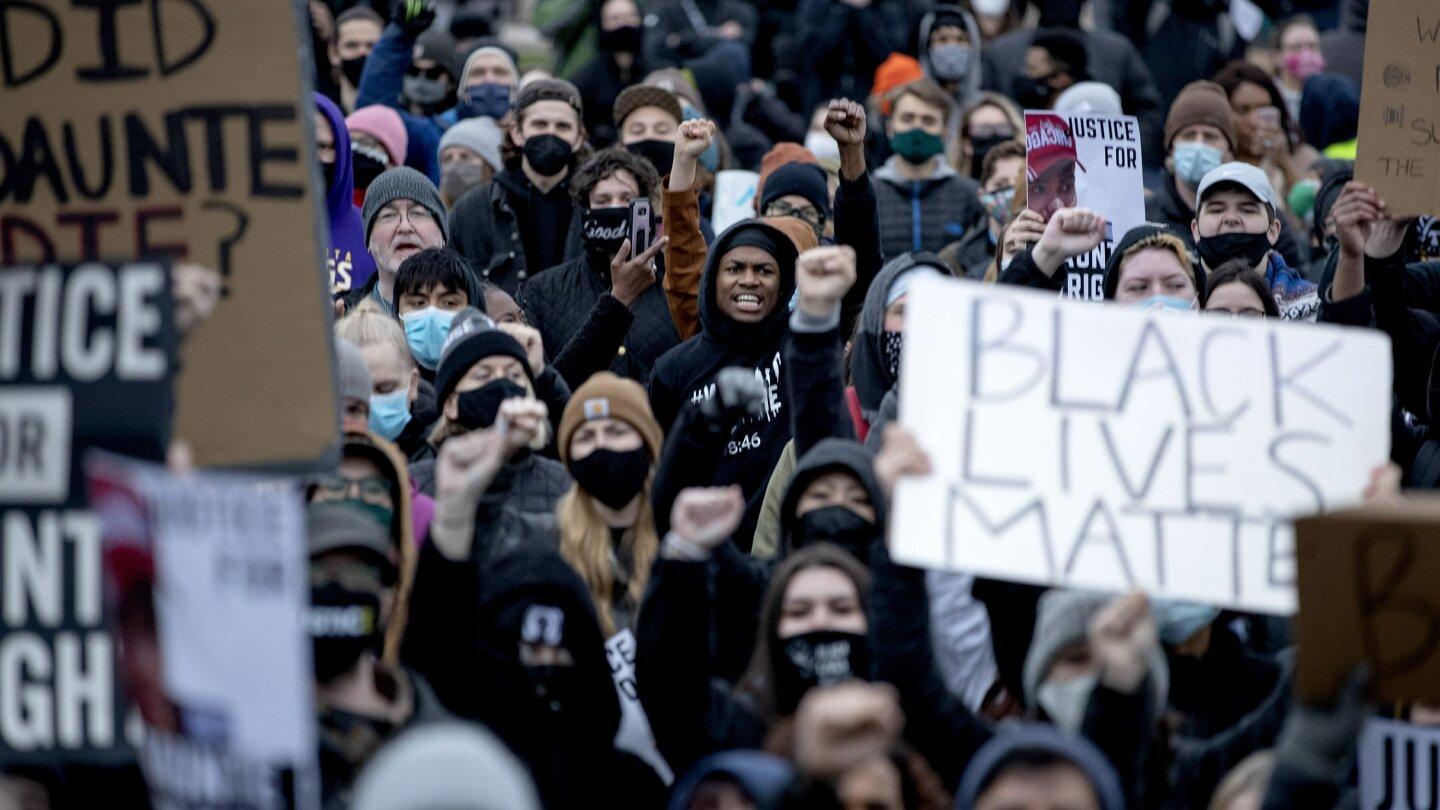} \\

    \begin{tikzpicture}[x=\linewidth, y=0.42cm]
        \def\nameright{0.12} \def\barleft{0.14} \def\barright{0.64} \def\barrange{0.50} \def\numleft{0.55}

        \node[anchor=east, font=\tiny\bfseries] at (\nameright,2) {MMDFND};
        \pgfmathsetmacro{\lenA}{\barleft+0.2242*\barrange}
        \fill[gray!50] (\barleft,1.75) rectangle (\lenA, 2.25);
        \node[anchor=west, font=\tiny] at (\numleft,2) {0.2242};

        \node[anchor=east, font=\tiny\bfseries] at (\nameright,1.2) {MiMOE};
        \pgfmathsetmacro{\lenB}{\barleft+0.2631*\barrange}
        \fill[gray!50] (\barleft,0.95) rectangle (\lenB, 1.45);
        \node[anchor=west, font=\tiny] at (\numleft,1.2) {0.2631};

        \node[anchor=east, font=\tiny\bfseries] at (\nameright,0.4) {\textsc{EGMD}};
        \pgfmathsetmacro{\lenC}{\barleft+0.1131*\barrange}
        \fill[teal!70] (\barleft,0.15) rectangle (\lenC, 0.65);
        \node[anchor=west, font=\tiny] at (\numleft,0.4) {\textbf{0.1131}};

        \draw[thick, black!50] (\barleft,-0.15) -- (\barright,-0.15);
        \node[anchor=north, font=\tiny] at (\barleft,-0.2) {Real};
        \node[anchor=north, font=\tiny] at (\barright,-0.2) {Fake};
    \end{tikzpicture}
    &
    \begin{tikzpicture}[x=\linewidth, y=0.42cm]
        \def\nameright{0.12} \def\barleft{0.14} \def\barright{0.64} \def\barrange{0.50} \def\numleft{0.55}

        \node[anchor=east, font=\tiny\bfseries] at (\nameright,2) {MMDFND};
        \pgfmathsetmacro{\lenD}{\barleft+0.4948*\barrange}
        \fill[red!55] (\barleft,1.75) rectangle (\lenD, 2.25);
        \node[anchor=west, font=\tiny] at (\numleft,2) {0.4948};

        \node[anchor=east, font=\tiny\bfseries] at (\nameright,1.2) {MiMOE};
        \pgfmathsetmacro{\lenE}{\barleft+0.5864*\barrange}
        \fill[red!55] (\barleft,0.95) rectangle (\lenE, 1.45);
        \node[anchor=west, font=\tiny] at (\numleft,1.2) {0.5864};

        \node[anchor=east, font=\tiny\bfseries] at (\nameright,0.4) {\textsc{EGMD}};
        \pgfmathsetmacro{\lenF}{\barleft+0.7623*\barrange}
        \fill[teal!70] (\barleft,0.15) rectangle (\lenF, 0.65);
        \node[anchor=west, font=\tiny] at (\numleft,0.4) {\textbf{0.7623}};

        \draw[thick, black!50] (\barleft,-0.15) -- (\barright,-0.15);
        \node[anchor=north, font=\tiny] at (\barleft,-0.2) {Real};
        \node[anchor=north, font=\tiny] at (\barright,-0.2) {Fake};
    \end{tikzpicture}
    \\

    \end{tabular}

    \caption{\textbf{Illustration of Semantic Misalignment.} Motivated by recurring prediction errors observed in our error analysis, we construct a clear example by pairing the same Health text with (i) its matched image and (ii) an unrelated Politics image. Bars report $P(\mathrm{Fake})$, illustrating how model confidence changes under pronounced cross-modal conflict.}
    \label{fig:semantic_misalignment}
\end{figure}
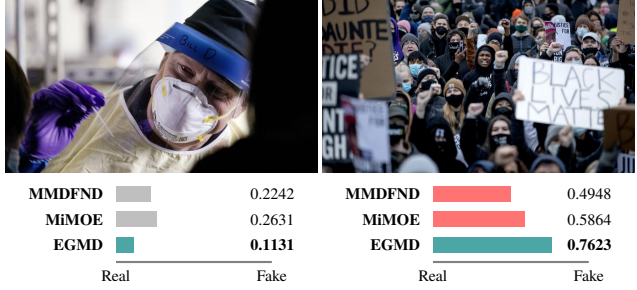
}

\section{Introduction}
\label{sec:intro}

The rapid proliferation of digital media has made fake news detection an indispensable task for safeguarding public trust~\citep{zeng2024mitigating, guo2025each}. Recently, multimodal learning has achieved superior performance in this domain, leveraging cross-modal correlations to significantly outperform unimodal counterparts~\citep{qi2021improving, ma2024event, yang2024reinforcement, liu2024teller, zhao2025unlearning}. Despite this progress, current methods exhibit poor cross-domain generalization, primarily due to two factors: \emph{domain bias} and \emph{semantic misalignment}. Domain bias typically originates from imbalanced data distributions and occurs when models rely on spurious domain-specific correlations rather than learning event-invariant semantics~\cite{li2024dual}. Moreover, multimodal frameworks encounter a unique challenge absent in unimodal settings known as semantic misalignment~\cite{cai2023exploring}. This issue stems from intrinsic cross-modal heterogeneity and manifests as semantic conflicts between modalities, significantly undermining the reliability of the detection process. Therefore, jointly addressing domain bias and semantic misalignment is essential for reliable cross-domain fake news detection.

However, the precise impact of these impediments on generalization remains underexplored. To disentangle and quantify these failure modes, we conducted a targeted empirical study to provide concrete evidence of these deficiencies:

\ding{182} \textbf{Domain Bias via Imbalanced Data Distributions.} Domain imbalance causes aggregate training objectives to be dominated by data-rich domains, encouraging models to learn majority-domain shortcuts while leaving decision boundaries for tail domains insufficiently estimated. This issue is particularly pronounced in Weibo21~\cite{nan2021mdfend}, where the \textit{Society} domain contains 2,615 instances, compared with only 236 in \textit{Science}, resulting in an 11.08$\times$ volume gap. Consistent with this imbalance, MMDFND~\cite{tong2024mmdfnd} and MiMOE-FND~\cite{liu2025modality} perform competitively in high-resource domains but exhibit substantial accuracy degradation in underrepresented domains such as \textit{Military} and \textit{Science}. Complete domain statistics and additional analysis are provided in the Supplementary Material.

\afterpage{\semanticmisalignmentfigure}

\ding{183} \textbf{Susceptibility to Semantic Misalignment.} Our error analysis revealed recurring prediction instability when textual and visual evidence was semantically inconsistent. To make this failure mode explicit, Figure~\ref{fig:semantic_misalignment} reconstructs a pronounced example by retaining the same Health-domain text and replacing its matched image with an unrelated Politics image. The figure reports changes in $P(\mathrm{Fake})$ rather than classification correctness: MMDFND and MiMOE-FND move toward the decision boundary (0.4948 and 0.5864), whereas \textsc{EGMD} exhibits a clearer response to the introduced conflict (0.7623). This reconstruction illustrates model sensitivity under deliberate semantic conflict, but does not establish that mismatch alone determines veracity. In practice, legitimate news may use stock or weakly related images; semantic mismatch is therefore neither a necessary nor a sufficient condition for fake news. A broader construct-validity evaluation would require separate annotations for deceptive mismatch and benign visual relevance.
Within this illustrative setting, the weaker response of existing models is consistent with a potential generalization bottleneck: cross-modal compatibility is introduced only \emph{after} fusion, when conflicting signals may already have contaminated the joint representation. This observation motivates our input-level calibration design, while the main performance claims remain grounded in the naturally labeled benchmark evaluations rather than this constructed example.

In this paper, we identify a shared error-propagation path: unreliable cross-modal evidence contaminates representations, entangles with domain statistics, and biases decisions. We therefore propose Expert-Guided Mutual Distillation (\textsc{EGMD}) with three dependent innovations. \emph{Input-Level Calibration} encodes pair-level coherence as a shared gain before fusion. \emph{Representation-Level Debiasing} constructs an Expert-Guided Teacher whose Dynamic Domain-specific Normalization (DDN) aligns domain-conditioned statistics before the MoE organizes residual domain patterns across experts. \emph{Decision-Level Generalization} builds a Mutual Distillation Student that combines domain prototypes and mutual learning with feature- and logit-level transfer. Each stage thus refines the preceding output rather than acting as an additive regularizer. The teacher-side modules are training-only, leaving one lightweight student branch for inference.

Our main contributions are summarized as follows:
\begin{itemize}
\item We reveal that domain imbalance and cross-modal semantic misalignment jointly encourage multimodal detectors to learn domain-specific shortcuts, leading to uneven performance across domains.
\item We propose \textsc{EGMD} for cross-domain uniformity. Its domain-normalized teacher aligns domain statistics and encourages domain-specific patterns to concentrate in specialized experts, while prototype-anchored mutual distillation discourages local priors.
\item We isolate how domain imbalance contributes to domain bias by constructing \textsc{Weibo\_Balanced}, which reduces the imbalance ratio from $11.08\times$ to $1.27\times$. Under this control, \textsc{EGMD}'s \textit{Total} falls by 37.7\%, supporting domain imbalance as an important driver of domain bias.
\item Experiments on four datasets confirm consistent gains, especially in vulnerable domains.
\end{itemize}

\section{Related Work}
\label{sec:related_work}

\subsection{Multimodal Fake News Detection.}
Multimodal technology has gained widespread recognition across diverse domains\cite{cui2025diffusion,cui2025multi,li2025srkd,li2025fedkd,li2025comae,li2025frequency},with applications extending to fake news detection\cite{qiao2025improving,shen2025gamed,zhang2025knowledge}.While early works focused on unimodal text analysis~\cite{dou2021user, zhu2022generalizing}, recent research has shifted toward multimodal approaches that leverage both textual and visual information\cite{wang2024bilateral,zhou2023multi}. State-of-the-art methods predominantly focus on improving cross-modal fusion mechanisms. For instance, COOLANT~\cite{wang2023cross} utilizes contrastive learning for alignment, FSRU~\cite{lao2024frequency} explores frequency domain interactions, and MTS~\cite{sun2025multimodal} captures high-order correlations via Taylor series expansion. Despite their success in extracting joint representations, these methods often overlook the critical issue of \textit{domain bias}, assuming independent and identically distributed (i.i.d.) data, which compromises their generalization to unseen domains.

\subsection{Domain Adaptation and MoE Architectures.}
Recent work addresses domain shift with Mixture-of-Experts (MoE) and distillation. DTDBD~\cite{li2024dual} uses dual-teacher distillation but is limited to unimodal inputs. In multimodal detection, BMR~\cite{ying2023bootstrapping}, MiMoE-FND~\cite{liu2025modality}, and MMDFND~\cite{tong2024mmdfnd} use experts for representation fusion, while MemiMoE-FND~\cite{meng2026memimoe} adds memory slots to hierarchical experts. ADOSE~\cite{chen2026adose} instead performs active multi-source adaptation by labeling a target-domain subset. These methods focus on fusion or supervised adaptation; \textsc{EGMD} jointly targets domain-level error disparity and semantic misalignment without target-domain veracity labels, using training-time expert guidance and mutual distillation.

A broader review of traditional and multimodal methods is provided in the Supplementary Material.

\section{Problem Formulation and Dataset Construction}
\label{sec:preliminaries}

\subsection{Problem Formulation}
We study cross-domain multimodal fake news detection under imbalanced domain distributions. The training set $\mathcal{D}=\bigcup_{d=1}^{K}\mathcal{D}_d$ covers $K$ domains with unequal sample counts $n_d=|\mathcal{D}_d|$. Each news item $\mathbf{x}=(T,V)\in\mathcal{X}=\mathcal{T}\times\mathcal{V}$ contains text $T$ and an image $V$, has a binary veracity label $y\in\{0,1\}$, where $y=1$ denotes fake news, and belongs to a domain $d\in\{1,\ldots,K\}$. The detector $f_{\theta}:\mathcal{X}\rightarrow[0,1]$ estimates the probability that an item is fake.

Let $P_d$ denote the data distribution of domain $d$ and $\ell$ the binary classification loss. To give every domain equal importance, our task-level objective targets the uniformly weighted domain risk
\begin{equation}
\min_{\theta}\;\frac{1}{K}\sum_{d=1}^{K}
\mathbb{E}_{(\mathbf{x},y)\sim P_d}
\left[\ell\!\left(f_{\theta}(\mathbf{x}),y\right)\right],
\label{eq:objective}
\end{equation}
which prevents performance in data-rich domains from defining the objective alone. Accordingly, the goal is not only high aggregate accuracy, but also strong tail-domain performance and low cross-domain error disparity.

\begin{table}[t]
\caption{Domain statistics before and after balancing. R and F denote real and fake instances, respectively.}
\centering
\footnotesize
\setlength{\tabcolsep}{2.5pt}
\renewcommand{\arraystretch}{1.0}
    \begin{tabular}{l ccc c ccc}
    \toprule
    \multirow{2.5}{*}{\textbf{Domain}} & \multicolumn{3}{c}{\textbf{Original}} & & \multicolumn{3}{c}{\textbf{Balanced (Ours)}} \\
    \cmidrule(lr){2-4} \cmidrule(lr){6-8}
    & R & F & \textbf{Size} & & R & F & \textbf{Size} \\
    \midrule
    
    Finance & 959 & 362 & 1,321 & & 1,033 & 430 & 1,463 \\
    Health & 485 & 515 & 1,000 & & 662 & 862 & 1,524 \\
    Military & 117 & 222 & 339 & & 1,057 & 325 & 1,382 \\
    Science & 143 & 93 & 236 & & 1,023 & 259 & 1,282 \\
    Politics & 304 & 546 & 850 & & 792 & 822 & 1,614 \\
    Disaster & 184 & 591 & 775 & & 721 & 867 & 1,588 \\
    Edu. & 243 & 248 & 491 & & 928 & 545 & 1,473 \\
    Society & 1,198 & 1,417 & 2,615 & & 714 & 913 & 1,627 \\
    Ent. & 1,007 & 440 & 1,447 & & 1,007 & 440 & 1,447 \\
    \midrule
    
    Sum & 4,640 & 4,434 & 9,074 & & 7,937 & 5,463 & 13,400 \\
    \midrule
    
    \multicolumn{8}{c}{Imbalance Ratio (Max/Min):} \\
    \multicolumn{8}{c}{
        Original: \textcolor{red}{\textbf{11.08$\times$}} \quad vs. \quad Ours: \textcolor{green!50!black}{\textbf{1.27$\times$}}
    } \\
    \bottomrule
    \end{tabular}
\label{tab:dataset_comparison_single}
\end{table}

\subsection{\textsc{Weibo\_Balanced}: Examining the Effect of Domain Imbalance}
\label{sec:benchmark}

To examine whether unequal domain volumes contribute to cross-domain performance disparities, we construct \textsc{Weibo\_Balanced}, a more evenly distributed extension of Weibo21. It retains the original task and nine-domain taxonomy while making the number of instances across domains comparable. This setting provides a complementary benchmark for evaluating whether domain disparities persist after the long-tailed distribution is substantially reduced. Table~\ref{tab:dataset_comparison_single} summarizes the resulting statistics.

\paragraph{Dataset Construction} We expand underrepresented domains through targeted keyword-based collection and reduce overrepresented domains through stratified downsampling. Following the Weibo21 collection protocol~\cite{nan2021mdfend}, ten domain experts independently annotate each newly collected instance. A domain label is retained when at least eight annotators agree; cases with lower agreement undergo further review.

\paragraph{Resulting Distribution} The resulting dataset contains 13,400 instances, with domain sizes ranging from 1,282 to 1,627. The maximum-to-minimum domain-size ratio decreases from $11.08\times$ in Weibo21 to $1.27\times$. This substantially reduces variation in domain volume and enables a more direct examination of its relationship with domain-level performance disparities.

\section{Methodology}
\label{sec:methodology}

\begin{figure*}[t!]
\centering
\includegraphics[width=1.0\textwidth]{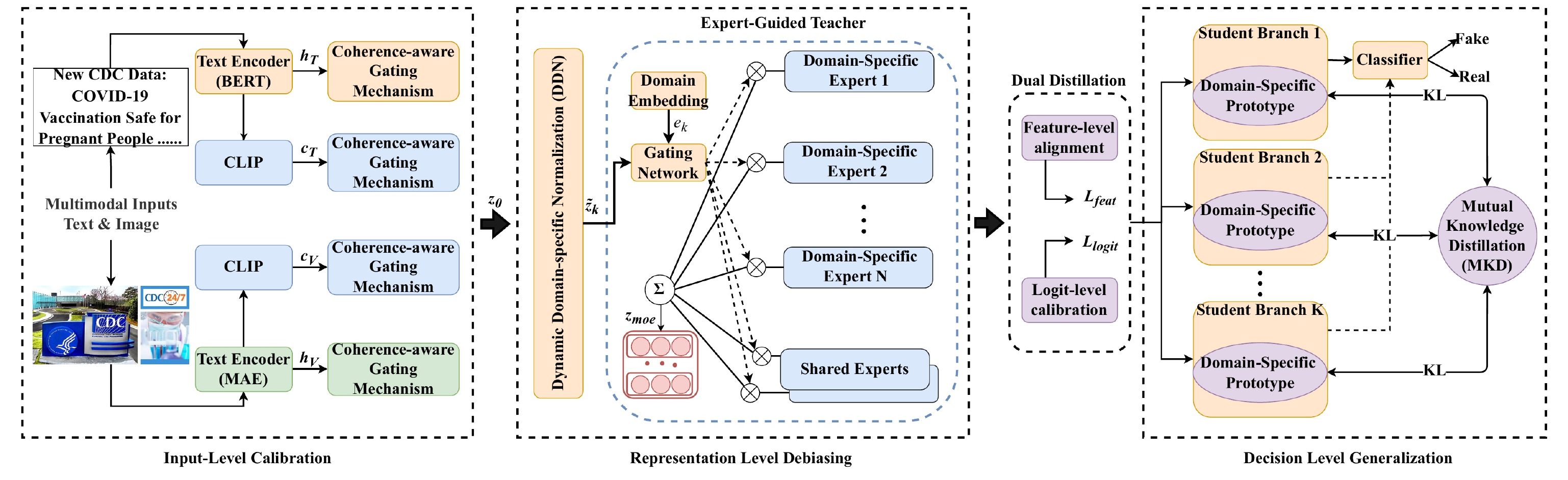} 
\caption{The \textsc{EGMD} framework comprises three stages: Input-Level Calibration (\S\ref{sec:input_calibration}), Representation-Level Debiasing via Dynamic Domain-specific Normalization and Mixture-of-Experts (\S\ref{sec:teacher}), and Decision-Level Generalization via a prototype-anchored Mutual Distillation Student (\S\ref{sec:student}). Teacher-student transfer uses feature- and logit-level distillation losses ($\mathcal{L}_{\text{feat}}$, $\mathcal{L}_{\text{logit}}$).}
\label{fig:framework}
\end{figure*}

Figure~\ref{fig:framework} presents \textsc{EGMD} as a three-stage pipeline. Input-Level Calibration exposes pairwise compatibility before fusion (\S\ref{sec:input_calibration}); Representation-Level Debiasing aligns domain statistics and organizes residual domain patterns through conditional experts (\S\ref{sec:teacher}); and Decision-Level Generalization transfers teacher knowledge through prototype-anchored mutual and dual-channel distillation (\S\ref{sec:student}).

\subsection{Input-Level Calibration}
\label{sec:input_calibration}

The first stage exposes cross-modal compatibility to the fusion network before feature interaction. Standard multimodal fusion presupposes that text and image convey coherent semantics, yet this assumption is frequently violated in cross-domain settings: a health-related headline may be paired with an unrelated political photograph, or a manipulated image may accompany factual text. Fusion without an explicit compatibility cue cannot distinguish these cases before their features interact.

We introduce a coherence-aware gating mechanism before feature interaction. Content representations $\mathbf{h}_{\mathcal{T}}\in\mathbb{R}^{d_{\mathcal{T}}}$ and $\mathbf{h}_{\mathcal{V}}\in\mathbb{R}^{d_{\mathcal{V}}}$ are extracted by BERT~\cite{devlin2019bert} and MAE~\cite{he2022masked}, while $\ell_2$-normalized CLIP references $\mathbf{c}_{\mathcal{T}},\mathbf{c}_{\mathcal{V}}\in\mathbb{R}^{d_c}$ provide a shared semantic coordinate system~\cite{radford2021learning}. Their pair-level compatibility is the scalar cosine similarity $\rho=\mathbf{c}_{\mathcal{T}}^{\top}\mathbf{c}_{\mathcal{V}}\in[-1,1]$. We map it to a sample-dependent shared gain $s\in(0,1)$:
\begin{equation}
    s = \operatorname{sigm}\!\left( w_s\,\rho/\tau_c + b_s \right),
    \label{eq:synergy}
\end{equation}
where $\tau_c>0$ is the coherence temperature and $w_s,b_s\in\mathbb{R}$ are learnable scalars. The conditioned representation is
\begin{equation}
    \mathbf{z}_0 = \mathcal{F}_{\text{fuse}}\!\left(
    \operatorname{concat}\!\left((1{+}s)\mathbf{h}_{\mathcal{T}},
    (1{+}s)\mathbf{h}_{\mathcal{V}}\right)\right),
    \label{eq:z0}
\end{equation}

where $\operatorname{concat}(\cdot,\cdot)\in\mathbb{R}^{d_{\mathcal{T}}+d_{\mathcal{V}}}$ and $\mathcal{F}_{\text{fuse}}:\mathbb{R}^{d_{\mathcal{T}}+d_{\mathcal{V}}}\rightarrow\mathbb{R}^{d_z}$. Equation~\ref{eq:z0} applies the same gain to both modalities. It therefore conditions the pair as a whole and does not identify, suppress, or reweight one modality relative to the other. In particular, $1+s\in(1,2)$ is strictly amplificatory: coherent pairs receive a larger gain, whereas low-coherence pairs remain near their original magnitude rather than being attenuated. We adopt this conservative operation because low CLIP similarity can also arise from benign abstraction or stock imagery. Accordingly, this component should be interpreted as an explicit compatibility cue, not as a modality-denoising gate; representation debiasing is handled by the subsequent teacher. Unlike the post-fusion similarity weighting in MMDFND~\citep{tong2024mmdfnd}, the compatibility signal enters our fusion network with the unimodal features.

\subsection{Representation-Level Debiasing: Expert-Guided Teacher}
\label{sec:teacher}

Given the compatibility-conditioned representation $\mathbf{z}_0$, the second stage addresses domain-dependent variation in feature statistics. Under imbalanced training data, a single shared transformation can be dominated by head domains and preserve correlations that transfer poorly to tail domains. The Expert-Guided Teacher therefore combines two operations. DDN standardizes batch-level feature statistics and applies domain-conditioned affine parameters, reducing raw statistical variation before routing. A domain-conditioned MoE then models the remaining heterogeneity through conditional expert specialization. Together, these operations produce the teacher representation used for decision-level distillation.

\paragraph{Step 1: Aligning Domain Distributions via DDN}
The first manifestation of domain bias is variation in feature statistics across domains, which can make a shared representation favor data-rich domains. We address this through Dynamic Domain-specific Normalization (DDN), which reduces differences in batch-level feature statistics before expert routing.

Concretely, we standardize $\mathbf{z}_0$ using batch-level statistics $(\boldsymbol{\mu}_{\mathcal{B}},\, \boldsymbol{\sigma}_{\mathcal{B}})$ and recover domain-appropriate structure via a learnable affine transformation $(\boldsymbol{\gamma}_k,\, \boldsymbol{\beta}_k)$ conditioned on domain identity $k$:
\begin{equation}
    \tilde{\mathbf{z}}_{k} = \boldsymbol{\gamma}_k \odot \frac{\mathbf{z}_0 - \boldsymbol{\mu}_{\mathcal{B}}}{\sqrt{\boldsymbol{\sigma}_{\mathcal{B}}^2 + \epsilon}} + \boldsymbol{\beta}_k,
    \label{eq:ddn}
\end{equation}
where $\epsilon$ is a small constant added for numerical stability. DDN serves a different purpose from the AdaIN module in MMDFND~\citep{tong2024mmdfnd}. AdaIN reweights the relative importance of cross-domain versus domain-specific knowledge after extraction; DDN operates proactively, standardizing the feature space before expert routing so that the subsequent MoE can focus on conditional specialization rather than compensating for distributional mismatch.

DDN removes a gain exactly only when that gain is constant across the entire mini-batch. Here $s_i$ is sample dependent. Even under the simplifying assumption that $\mathcal{F}_{\mathrm{fuse}}$ is positively homogeneous, so that $\mathbf{z}_{0,i}=a_i\mathbf{u}_i$ with $a_i=1+s_i$, normalization uses $\boldsymbol{\mu}_{\mathcal{B}}=|\mathcal{B}|^{-1}\sum_j a_j\mathbf{u}_j$ and the corresponding feature-wise variance. Hence $a_i$ cannot be factored out unless all $a_i$ are equal. DDN may nevertheless attenuate magnitude information, so we do not claim that the gain is preserved unchanged; it provides lightweight pre-fusion conditioning, whereas DDN targets batch-level domain statistics.

\paragraph{Step 2: Concentrating Domain-Specific Patterns via MoE}
Even following distributional alignment, the representation may continue to encode domain-specific shortcuts, which are statistical patterns that correlate with veracity labels within a single domain but fail to generalize. We employ a MoE layer with domain-conditioned routing to encourage these patterns to concentrate in specialized experts while allowing shared experts to model recurring veracity cues.

The gating network $\mathcal{G}$ routes each sample based on both the aligned feature $\tilde{\mathbf{z}}_k\in\mathbb{R}^{d_z}$ and an explicit learnable domain embedding $\mathbf{e}_k \in \mathbb{R}^{d_e}$:
\begin{equation}
    \boldsymbol{\alpha} = \mathrm{Softmax}\!\left( \mathbf{W}_g \cdot \mathrm{ReLU}\!\left([\tilde{\mathbf{z}}_{k} \;;\; \mathbf{e}_k]\right) \right),
    \label{eq:gating}
\end{equation}
where $N_e$ is the number of experts and $\mathbf{W}_g\in\mathbb{R}^{N_e\times(d_z+d_e)}$ is the learned weight matrix of the gating network. The teacher's output is the expert-weighted aggregate:
\begin{equation}
    \mathbf{z}_{\text{moe}} = \textstyle\sum_{i} \alpha_i\, E_i(\tilde{\mathbf{z}}_{k}),
    \label{eq:moe}
\end{equation}
where each expert $E_i$ is a feed-forward network. Through domain-conditioned routing, the weighted aggregation encourages domain-specific patterns to concentrate in specialized experts, while shared experts model recurring veracity signals.

\subsection{Decision-Level Generalization: Mutual Distillation Student}
\label{sec:student}

The final stage transfers the teacher's knowledge to the student's decision process. Even with the teacher representation, a student trained via standard supervision can still overfit to domain-specific priors if its decision boundaries are not explicitly constrained. We address this through two complementary mechanisms, namely geometric anchoring and mutual consensus, and connect the teacher to the student via dual-channel distillation.

\paragraph{Geometric Anchoring via Domain Prototypes}
Without explicit constraints, domain-specific student branches may diverge into isolated clusters. To prevent this, we introduce domain-specific prototype memories. For each domain $k$, a prototype $\mathbf{p}_k$ tracks the moving centroid of the feature distribution:
\begin{equation}
    \mathbf{p}_{k} \leftarrow \lambda\, \mathbf{p}_{k} + (1 - \lambda) \cdot \mathbb{E}_{\mathbf{h} \in \mathcal{B}_k}[\mathbf{h}],
    \label{eq:prototype}
\end{equation}
where $\lambda$ is a momentum coefficient and $\mathcal{B}_k$ denotes the mini-batch samples from domain $k$. These prototypes function as geometric anchors: they exert a centripetal force on each branch's feature distribution, implicitly restricting manifold divergence.

\paragraph{Enforcing Cross-Domain Consensus via Mutual Distillation}
Beyond geometric anchoring, we further smooth the decision surface through mutual knowledge distillation. Each student branch $S_k$ specializes in domain $k$, but we require all branches to reach a consensus on veracity prediction. We enforce this by minimizing the pairwise KL divergence between branches:
\begin{equation}
    \mathcal{L}_{\text{mutual}} = \sum_{k \neq j} \mathrm{KL}\!\left( P(Y|S_k) \,\|\, P(Y|S_j) \right).
    \label{eq:mkd}
\end{equation}

\paragraph{Bridging Teacher and Student via Dual Distillation}
We close the gap between the teacher and student through dual-channel distillation that transfers knowledge at two complementary levels. First, feature-level alignment synchronizes the student's feature topology with the teacher's representation space via an InfoNCE-based contrastive objective:
\begin{equation}
    \mathcal{L}_{\text{feat}} = -\log \frac{\exp\!\left(\mathrm{sim}(\mathbf{z}_s, \mathbf{z}_t) / \tau_f\right)}{\sum_{j}\exp\!\left(\mathrm{sim}(\mathbf{z}_s, \mathbf{z}_j) / \tau_f\right)},
    \label{eq:feat}
\end{equation}
where $\mathbf{z}_s$ and $\mathbf{z}_t$ denote student and teacher features for the same instance, and $\mathbf{z}_j$ are negative samples within the batch. Second, logit-level calibration distills the teacher's soft probability distribution into the student:
\begin{equation}
    \mathcal{L}_{\text{logit}} = \mathrm{KL}\!\left( \operatorname{softmax}(\boldsymbol{\ell}_t / T_d) \,\|\, \operatorname{softmax}(\boldsymbol{\ell}_s / T_d) \right),
    \label{eq:logit}
\end{equation}
where $\boldsymbol{\ell}_t, \boldsymbol{\ell}_s\in\mathbb{R}^{2}$ are the teacher and student logits for the two veracity classes, and $T_d>0$ is the logit-distillation temperature.

\subsection{Joint Optimization}
\label{sec:optimization}

The entire framework is optimized end-to-end with a composite objective:
\begin{equation}
    \mathcal{L}_{\text{total}} = \underbrace{\mathcal{L}_{\text{CE}}}_{\text{veracity}} + \;\alpha\, \underbrace{\mathcal{L}_{\text{mutual}}}_{\text{consensus}} + \;\beta\, \underbrace{(\mathcal{L}_{\text{logit}} + \mathcal{L}_{\text{feat}})}_{\text{distillation}},
    \label{eq:total}
\end{equation}
where $\mathcal{L}_{\text{CE}}$ is the standard binary cross-entropy loss for veracity classification, and $\alpha, \beta$ balance the contributions of mutual distillation and dual-channel distillation, respectively.

\paragraph{Training and Inference}
\textsc{EGMD} is jointly optimized using $\mathcal{L}_{\text{total}}$, with training-time domain labels routing instances to domain-specific components. When domain labels are unavailable at inference, prototype similarity selects one source-domain student branch. Detailed training settings and routing robustness are provided in the Supplementary Material.

\begin{figure}[t]
\centering
\includegraphics[width=\columnwidth]{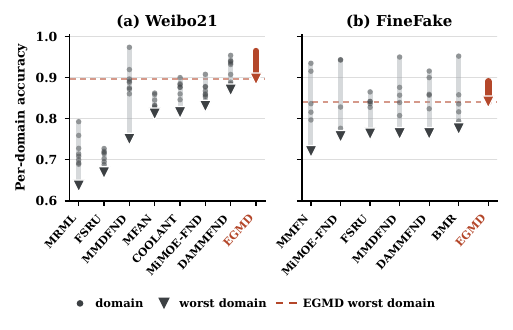}
\caption{Per-domain accuracy ranges on Weibo21 and FineFake. Triangles mark worst-domain accuracy; dashed lines indicate the \textsc{EGMD} lower bound.}
\label{fig:domain_distribution}
\end{figure}

\begin{table*}[t]
\caption{Performance comparison on Chinese datasets. Worst Acc. denotes the minimum domain accuracy, and Total is FNED+FPED. The proposed \textbf{\textsc{EGMD}} is highlighted in light gray.}
\renewcommand{\arraystretch}{1.0}
\centering
\footnotesize 
\setlength{\tabcolsep}{1.2pt}

    \begin{tabular}{l c c c c c c c c c c c c c}
    \toprule
    \multirow{2.5}{*}{Method}
    & \multirow{2.5}{*}{Finance} & \multirow{2.5}{*}{Health} & \multirow{2.5}{*}{Military} & \multirow{2.5}{*}{Science} & \multirow{2.5}{*}{Politics} & \multirow{2.5}{*}{Int./Dis.} & \multirow{2.5}{*}{Education} & \multirow{2.5}{*}{Society} & \multirow{2.5}{*}{Ent.}
    & \multicolumn{4}{c}{Overall} \\
    \cmidrule(lr){11-14}
    & & & & & & & & & & Accuracy $\uparrow$ & Worst Acc. $\uparrow$ & Total $\downarrow$ & PFS $\uparrow$\\
    \midrule

    \multicolumn{14}{c}{\cellcolor{gray!15}\textbf{Dataset: Weibo}} \\
    MFAN & 0.7949 & 0.8780 & 0.8633 & 0.8090 & 0.7774 & 0.8325 & 0.8192 & 0.8599 & 0.8734 & 0.8552 & 0.7774 & 1.9217 & 0.6884\\
    MRML & 0.7889 & 0.9212 & 0.8222 & 0.7397 & 0.8140 & 0.8846 & 0.6364 & 0.7692 & 0.8284 & 0.7930 & 0.6364 & 1.1009 & 0.6565\\
    COOLANT & 0.8275 & 0.8333 & \textbf{0.9390} & 0.8538 & 0.8973 & 0.8833 & 0.8520 & 0.8372 & 0.8602 & 0.8490 & 0.8275 & \underline{0.8117} & 0.7186\\
    FSRU & 0.8548 & 0.9162 & 0.8810 & 0.8537 & 0.8564 & 0.7444 & 0.8792 & \textbf{0.9588} & 0.9197 & \underline{0.9265} & 0.7444 & 1.1888 & 0.7598\\
    MMDFND & 0.8571 & 0.9052 & 0.8696 & 0.7885 & 0.7353 & \textbf{0.9725} & 0.9250 & \underline{0.9393} & 0.9216 & 0.9147 & 0.7353 & 1.8703 & 0.7365\\
    MiMOE-FND & \textbf{0.9033} & \textbf{0.9306} & 0.9108 & 0.8930 & \underline{0.9297} & \underline{0.8953} & \underline{0.9306} & 0.9072 & 0.8772 & 0.9045 & \underline{0.8772} & 0.8406 & 0.7608\\
    DAMMFND & 0.8163 & 0.8966 & 0.8696 & \underline{0.9231} & 0.8235 & 0.8889 & 0.9250 & \underline{0.9393} & \textbf{0.9477} & 0.9205 & 0.8163 & 0.9031 & \underline{0.7693}\\
    \rowcolor{black!10} \textbf{\textsc{EGMD}} & \underline{0.8980} & \underline{0.9224} & \underline{0.9200} & \textbf{0.9423} & \textbf{0.9412} & 0.8889 & \textbf{0.9625} & 0.9347 & \underline{0.9434} & \textbf{0.9348\textsuperscript{†}} & \textbf{0.8889} & \textbf{0.5511\textsuperscript{†}} & \textbf{0.8143\textsuperscript{†}} \\

    \addlinespace[5pt]
    \multicolumn{14}{c}{\cellcolor{gray!15}\textbf{Dataset: Weibo21}} \\
    MFAN & 0.8156 & 0.8454 & 0.8142 & 0.8114 & 0.8624 & 0.8590 & 0.8325 & 0.8312 & 0.8271 & 0.8341 & 0.8114 & 1.0826 & 0.6902\\
    MRML & 0.6892 & 0.7000 & 0.6923 & 0.7592 & 0.7143 & 0.7279 & 0.6364 & 0.7086 & 0.7928 & 0.7172 & 0.6364 & \underline{0.5632} & 0.6386\\
    COOLANT & 0.8768 & 0.8148 & 0.9000 & 0.8827 & 0.8760 & 0.8462 & 0.8853 & 0.8230 & 0.8602 & 0.8506 & 0.8148 & 0.9965 & 0.7077\\
    FSRU & 0.6952 & 0.6883 & 0.6686 & 0.6750 & 0.7151 & 0.7176 & 0.7206 & 0.7275 & 0.7030 & 0.7086 & 0.6686 & 0.9454 & 0.5971\\
    MMDFND & \underline{0.8913} & 0.8879 & \textbf{0.9737} & 0.7500 & 0.8750 & \underline{0.8966} & 0.8605 & 0.8720 & \underline{0.9195} & 0.8865 & 0.7500 & 1.5860 & 0.7176\\
    MiMOE-FND & 0.8601 & 0.8576 & 0.8772 & 0.8303 & \underline{0.9080} & 0.8674 & 0.8535 & 0.8783 & 0.8432 & 0.8670 & 0.8303 & 0.9058 & 0.7263\\
    DAMMFND & 0.8846 & \underline{0.8889} & 0.8696 & \underline{0.9412} & \textbf{0.9394} & \textbf{0.9318} & \underline{0.9355} & \underline{0.9076} & \textbf{0.9541} & \underline{0.9158} & \underline{0.8696} & 0.9301 & \underline{0.7638}\\
    \rowcolor{black!10} \textbf{\textsc{EGMD}} & \textbf{0.9275} & \textbf{0.9138} & \underline{0.9474} & \textbf{0.9643} & 0.9038 & \underline{0.8966} & \textbf{0.9535} & \textbf{0.9080} & 0.9128 & \textbf{0.9163} & \textbf{0.8966} & \textbf{0.5415\textsuperscript{†}} & \textbf{0.8008\textsuperscript{†}}\\

    \bottomrule
    \end{tabular}
\label{tab:comparison}
\end{table*}

\begin{table*}[t]
\caption{Performance comparison on FineFake. Worst Acc. denotes the minimum domain accuracy, and Total is FNED+FPED.}

\renewcommand{\arraystretch}{1.0}
\centering
\footnotesize
\setlength{\tabcolsep}{3.5pt}

    \begin{tabular}{l c c c c c c c c c c}
    \toprule
    \multirow{2.5}{*}{Method} & \multirow{2.5}{*}{Business} & \multirow{2.5}{*}{Politics} & \multirow{2.5}{*}{Society} & \multirow{2.5}{*}{Health} & \multirow{2.5}{*}{Ent.} & \multirow{2.5}{*}{Conflict} & \multicolumn{4}{c}{Overall} \\
    \cmidrule(lr){8-11}
    & & & & & & & Accuracy $\uparrow$ & Worst Acc. $\uparrow$ & Total $\downarrow$ & PFS $\uparrow$\\
    \midrule

    \multicolumn{11}{c}{\cellcolor{gray!15}\textbf{Dataset: FineFake}} \\
    MRML & 0.6484 & 0.6089 & 0.5886 & 0.6263 & 0.6263 & -- & 0.6122 & -- & 0.7671 & 0.5329\\
    MMFN & \underline{0.9155} & 0.7969 & 0.8158 & 0.9348 & 0.8366 & 0.7203 & 0.8193 & 0.7203 & 0.9134 & 0.6876\\
    BMR & 0.8581 & 0.7753 & 0.8167 & \textbf{0.9522} & 0.8358 & 0.7938 & 0.8144 & \underline{0.7753} & 0.8053 & 0.6915\\
    FSRU & 0.8368 & \underline{0.8652} & 0.7629 & 0.8278 & 0.8421 & \underline{0.8418} & 0.8313 & 0.7629 & \underline{0.4979} & 0.7389\\
    MMDFND & 0.8571 & 0.8394 & 0.8077 & \underline{0.9500} & 0.8762 & 0.7636 & 0.8427 & 0.7636 & 1.0137 & 0.7005\\
    MiMOE-FND & \textbf{0.9426} & 0.7565 & 0.7767 & 0.9435 & 0.8284 & 0.7627 & 0.7994 & 0.7565 & 1.1675 & 0.6589\\
    DAMMFND & 0.8571 & 0.8594 & \underline{0.8242} & 0.9159 & \textbf{0.9000} & 0.7636 & \underline{0.8579} & 0.7636 & 0.6448 & \underline{0.7414}\\
    \rowcolor{black!10} \textbf{\textsc{EGMD}} & 0.8776 & \textbf{0.8835} & \textbf{0.8407} & 0.8800 & \underline{0.8810} & \textbf{0.8909} & \textbf{0.8733\textsuperscript{†}} & \textbf{0.8407} & \textbf{0.3603\textsuperscript{†}} & \textbf{0.7959\textsuperscript{†}} \\
    \bottomrule
    \end{tabular}
\label{tab:performance_comparison}
\end{table*}

\section{Experiment}

\subsection{Experiment Setup}

\textbf{Datasets.}
We evaluate on three public multimodal fake-news benchmarks: the Chinese \textbf{Weibo}~\cite{wang2018eann} and \textbf{Weibo21}~\cite{nan2021mdfend} datasets, and the English \textbf{FineFake}~\cite{zhou2026finefake} dataset. We additionally use the constructed \textbf{\textsc{Weibo\_Balanced}} benchmark (Section~\ref{sec:benchmark}) as a controlled setting for examining the effect of domain imbalance. Together, these four datasets cover two languages and diverse news domains.

\textbf{Baselines.}
We select \textbf{MFAN}~\cite{zheng2022mfan}, \textbf{MRML}~\cite{peng2023mrml}, \textbf{COOLANT}~\cite{wang2023cross}, \textbf{BMR}~\cite{ying2023bootstrapping}, \textbf{FSRU}~\cite{lao2024frequency}, \textbf{MMDFND}~\cite{tong2024mmdfnd}, \textbf{MiMOE-FND}~\cite{liu2025modality}, and \textbf{DAMMFND}~\cite{lu2025dammfnd} as baselines that can be evaluated under the same data splits and supervision protocol. The protocol differences of concurrent methods are documented in the Supplementary Material.

\textbf{Metrics.}
We report aggregate Accuracy, worst-domain accuracy $\text{Worst}=\min_{d\in\mathcal{D}}\text{Accuracy}_d$, and cross-domain error disparity $\text{Total}=\text{FNED}+\text{FPED}$~\cite{liu2021authors, li2024dual, feng2026c2po, dixon2018measuring}. We also provide the Performance Fairness Synergistic Score (PFS):
\begin{equation}
PFS = \theta \cdot \text{ACC} + (1-\theta) \cdot e^{-2 \cdot \text{Total}}
\label{eq:pfs}
\end{equation}
We fix $\theta=0.8$ to prioritize accuracy and use the exponential term as a bounded penalty for disparity. Because this scalarization is preference-dependent, PFS is auxiliary; our conclusions rely primarily on Accuracy, Worst, and \textit{Total}.

For a fair comparison, all baselines use identical settings. Results are averaged over five independent runs; a $t$-test against the runner-up determines significance ($p<0.05$), marked by \textsuperscript{\dag}. Full implementation, dataset, baseline, and metric details are provided in the Supplementary Material.

\subsection{Aggregate Accuracy Masks Domain Failures}
\label{sec:main_results}

Aggregate accuracy conceals substantial domain-level variation (Figure~\ref{fig:domain_distribution}; Tables~\ref{tab:comparison} and \ref{tab:performance_comparison}). On Weibo21, DAMMFND and \textsc{EGMD} appear nearly tied in average accuracy (91.58\% versus 91.63\%). Their least accurate domains, however, differ by 2.70 points (86.96\% versus 89.66\%), and their \textit{Total} disparities differ by 41.8\% (0.9301 versus 0.5415). MMDFND illustrates the masking effect more sharply: its 88.65\% average hides a 22.37-point range between \textit{Military} (97.37\%) and \textit{Science} (75.00\%). The pattern persists on FineFake, where DAMMFND averages 85.79\% but falls to 76.36\% in its weakest domain; \textsc{EGMD} raises these figures to 87.33\% and 84.07\%, respectively. Thus, a similar mean can conceal materially different deployment risk. We consequently interpret average Accuracy together with Worst Acc. and \textit{Total}, rather than treating the mean as the headline result.

\subsection{Tail-Domain Performance}
\label{sec:tail_analysis}

The per-domain breakdown locates where the improvement occurs. In the low-resource Weibo21 \textit{Science} domain, \textsc{EGMD} reaches 96.43\%, 2.31 points above the best competing result from DAMMFND. On FineFake \textit{Conflict}, it reaches 89.09\%, improving over FSRU's 84.18\% by 4.91 points. \textit{Military} provides an informative exception: MMDFND remains 2.63 points higher than \textsc{EGMD} (97.37\% versus 94.74\%). The result is therefore not a uniform gain in every domain. Rather, \textsc{EGMD} compresses the lower tail: it obtains the highest Worst Acc. on Weibo, Weibo21, and FineFake, exceeding the strongest competing worst-domain values by 1.17, 2.70, and 6.54 points, respectively. This concentration at the lower end explains why the small change in mean accuracy in Section~\ref{sec:main_results} corresponds to a substantially more uniform detector.

\begin{figure}[t]
\centering
\includegraphics[width=\columnwidth]{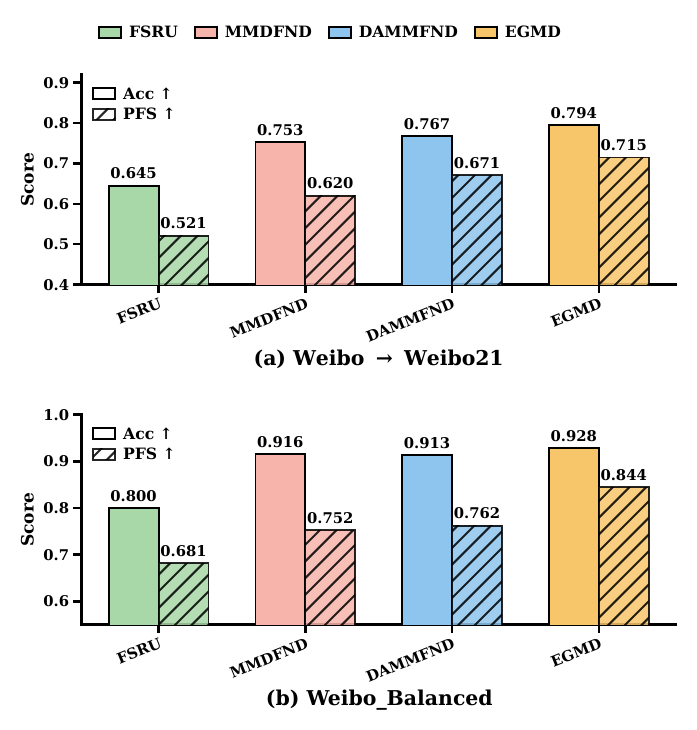}
\caption{Overall Accuracy and PFS on (a) cross-dataset generalization (train on Weibo, test on Weibo21) and (b) the \textsc{Weibo\_Balanced} benchmark. \textsc{EGMD} achieves the best Accuracy and PFS in both settings. Per-domain results are provided in the Supplementary Material.}
\label{fig:overall_combined}
\end{figure}

\subsection{Does Domain Imbalance Drive Domain Bias?}
\label{sec:balanced_analysis}

We use \textsc{Weibo\_Balanced} as a controlled intervention on domain volume, reducing the imbalance ratio from $11.08\times$ to $1.27\times$. Under this intervention (Figure~\ref{fig:overall_combined}b), \textsc{EGMD}'s accuracy rises from 91.63\% to 92.80\%, while \textit{Total} falls from 0.5415 to 0.3376, a further 37.7\% reduction. Balancing alone does not produce the same behavior in the strongest baseline: DAMMFND changes from 91.58\%/0.9301 to 91.34\%/0.9293 in Accuracy/\textit{Total}. MMDFND's disparity decreases from 1.5860 to 1.1511 but remains more than three times that of \textsc{EGMD}. These contrasts support domain imbalance as a major driver of the domain bias exposed in Section~\ref{sec:main_results}, while showing that balanced volume must interact with a model capable of preventing domain-specific shortcuts. The simultaneous improvement in Accuracy and \textit{Total} also shows that lower disparity is not obtained by sacrificing aggregate detection performance. More broadly, the differing responses across models indicate that domain volume is a contributing factor rather than a sufficient explanation; model design determines whether balanced evidence yields stable domain-wise decisions. The cross-dataset result in Figure~\ref{fig:overall_combined}a provides a complementary stress test: \textsc{EGMD} also retains the highest Accuracy and PFS when trained on Weibo and evaluated on Weibo21.

\subsection{What Reduces Domain Disparity?}

\begin{figure}[t]
\centering
\includegraphics[width=\columnwidth]{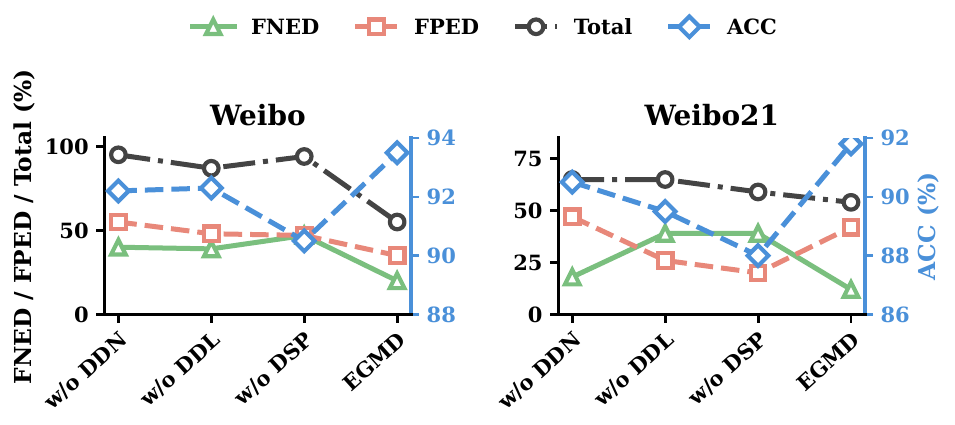}
\caption{Ablation study on Weibo and Weibo21. DDN: Dynamic Domain-specific Normalization; DDL: Dual Distillation Losses; DSP: Domain-Specific Prototypes.}
\label{fig:ablation}
\end{figure}

Figure~\ref{fig:ablation} evaluates two levels of \textsc{EGMD}. At the representation level, removing DDN raises \textit{Total} from 0.5511 to 0.9425 on Weibo and from about 0.54 to 0.65 on Weibo21, while accuracy falls by only 1.3 points in each case. This difference between disparity and accuracy indicates that statistical alignment redistributes performance across domains rather than merely increasing average capacity. At the decision level, removing dual-channel distillation increases disparity, whereas removing domain-specific prototypes yields the largest accuracy losses: 3.0 points on Weibo and 3.8 points on Weibo21. Together, representation-level debiasing reduces domain variation, and decision-level generalization converts aligned representations into accurate, stable predictions.

\subsection{Parameter Sensitivity}
\begin{figure}[t]
\centering
\includegraphics[width=\columnwidth]{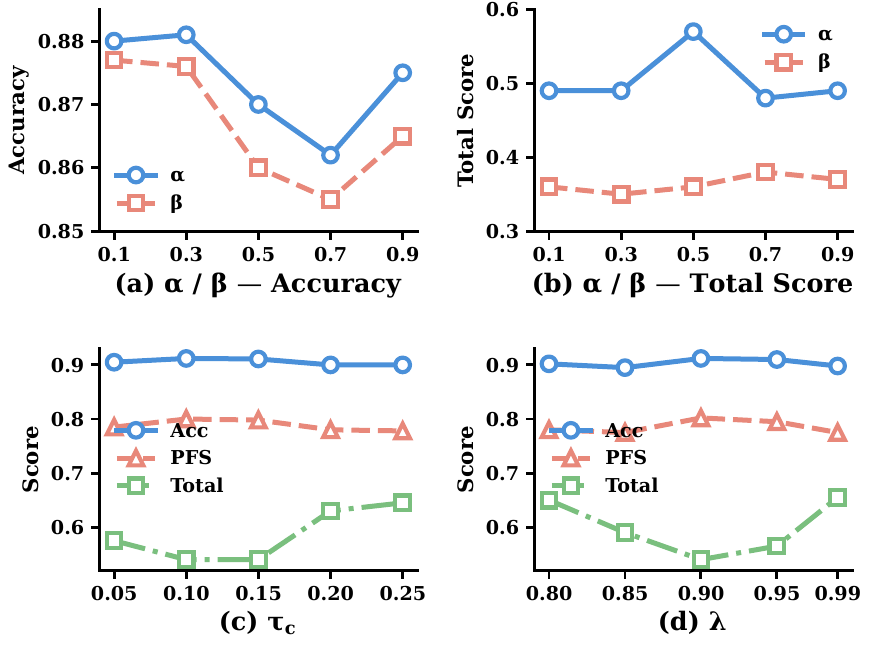}
\caption{Parameter sensitivity of $\alpha$, $\beta$, $\tau_c$, and $\lambda$ in terms of detection performance and cross-domain disparity.}
\label{fig:parameter_analysis}
\end{figure}

Figure~\ref{fig:parameter_analysis} examines the loss weights $\alpha$ and $\beta$, coherence temperature $\tau_c$, and prototype momentum $\lambda$; $\tau_f$ and $T_d$ remain fixed. The model is more sensitive to $\alpha$ than to $\beta$. Setting $\alpha=0.3$ gives the highest accuracy (0.881) with $\textit{Total}=0.49$, whereas $\alpha=0.5$ increases disparity to 0.57 without improving accuracy. By contrast, \textit{Total} varies only from 0.35 to 0.38 across $\beta$, indicating that the method is comparatively robust to this weight. The representation-level parameters show clearer optima. At $\tau_c=0.1$, the model reaches 0.912 Accuracy and 0.800 PFS with the lowest \textit{Total} of 0.54; larger temperatures weaken this balance. Similarly, $\lambda=0.9$ maximizes Accuracy and PFS while minimizing disparity, whereas values on either side increase \textit{Total}. We therefore use $\alpha=\beta=0.3$, $\tau_c=0.1$, and $\lambda=0.9$ throughout.

\FloatBarrier
\section{Conclusion}
We propose Expert-Guided Mutual Distillation (\textsc{EGMD}), a unified teacher--student framework for domain bias and semantic misalignment in multimodal fake news detection. Its key innovation is a dependent pipeline that conditions fusion on pair-level coherence, organizes domain-conditioned variation through a normalized expert teacher, and transfers feature geometry and predictions to lightweight students via prototype-anchored dual-channel distillation. \textsc{Weibo\_Balanced} further isolates the effect of data imbalance. Across four datasets, \textsc{EGMD} improves aggregate and tail-domain accuracy while achieving the lowest cross-domain error disparity.

\paragraph{Acknowledgments.}
This work was supported in part by the National Natural Science Foundation of China under Grant U22A2099 and Grant 62336003, and in part by the Fundamental Research Funds for the Central Universities under Grant 88022124.

\bibliography{sample-base}

\clearpage
\appendix
\begin{center}
{\LARGE\bfseries Appendix}
\end{center}
\noindent\textbf{Appendix organization.}
Sections~A--E provide reproducibility details and extended background, including implementation, datasets, baselines, metrics, and related work. Sections~F--J present supplementary evidence on domain bias, the construction of \textsc{Weibo\_Balanced}, efficiency and deployment robustness, cross-dataset generalization, and feature visualization.

\section{Implementation Details}
\label{app:implementation detail}

\subsection{Training Configuration}

\textbf{Environment and reproducibility.}
The proposed method is implemented in PyTorch, and all experiments are conducted on an NVIDIA GeForce RTX 4090 GPU. All reported results are averaged over five independent runs using five distinct, pre-specified random seeds. The same seed set is used for every method and dataset.

\textbf{Feature extraction.}
The text branch uses pretrained BERT~\cite{devlin2019bert} and CLIP~\cite{radford2021learning} text encoders. BERT produces a frozen 768-dimensional representation. The visual branch resizes each image to $224\times224$ pixels and extracts features with pretrained MAE and CLIP encoders. Multimodal features are projected through a single-layer MLP with a hidden dimension of 384 and dropout of 0.2.

\textbf{Optimization.}
We use an initial learning rate of $1\times10^{-4}$, a batch size of 64, and a maximum of 50 epochs. Training stops early when validation performance does not improve for five consecutive epochs.

\textbf{Joint training.}
The teacher and collaborative students are optimized jointly using the composite objective $\mathcal{L}_{\text{total}}$; the teacher receives no separate pretraining. The cross-entropy veracity loss is applied to both teacher and student predictions. During training, ground-truth domain labels route instances to the corresponding domain-specific experts and student branches, following the supervision available in standard multi-domain benchmarks.

\section{Dataset Details}
\label{sec:datasets}
We evaluate on three public multimodal benchmarks, Weibo~\cite{wang2018eann}, Weibo21~\cite{nan2021mdfend}, and FineFake~\cite{zhou2026finefake}, together with the constructed \textsc{Weibo\_Balanced} benchmark detailed in Appendix~\ref{app:dataset_details}. All datasets contain paired text-image news and binary veracity labels across multiple domains.

\textbf{FineFake}~\cite{zhou2026finefake} is a multimodal, knowledge-enhanced benchmark for cross-domain fake news detection, covering six themes across eight platforms. We exclude corrupted instances and report the retained subset in Table~\ref{tab:finefake_stats}.

\begin{table}[H]
\caption{Statistics of the FineFake dataset, where Bus., Pol., Soc., Heal., Ent., and Con. represent Business, Politics, Society, Health, Entertainment, and Conflict, respectively.}
\renewcommand{\arraystretch}{1.3}
\centering
\footnotesize
\setlength{\tabcolsep}{2.7pt}
\begin{tabular}{lrrrrrrr}
\hline
Domain & Bus. & Pol. & Soc. & Heal. & Ent. & Con. & All \\
\hline
Real & 77 & 704 & 646 & 162 & 513 & 155 & 2,257 \\
Fake & 219 & 1,054 & 505 & 198 & 827 & 199 & 3,002 \\
\hline
Total & 296 & 1,758 & 1,151 & 360 & 1,340 & 354 & 5,259 \\
\hline
\end{tabular}
\label{tab:finefake_stats}
\end{table}

\textbf{Weibo}~\cite{wang2018eann} is a Chinese multi-domain fake news detection dataset collected from Sina Weibo. It covers finance, health, military, science, politics, international, education, society, and entertainment. Table~\ref{tab:weibo_stats} reports its domain and label distribution.

\begin{table}[H]
\caption{Statistics of the Weibo dataset, where Int., Edu., Soc., and Ent. represent international, education, society, and entertainment, respectively.}
\renewcommand{\arraystretch}{1.0}
\centering
\small
\setlength{\tabcolsep}{3pt}
\begin{tabular}{>{\centering}p{1.05cm}>{\centering}p{1.05cm}>{\centering}p{1.05cm}>{\centering}p{1.05cm}>{\centering}p{1.05cm}>{\centering\arraybackslash}p{1.05cm}}
\hline
Domain & Finance & Health & Military & Science & Politics \\
\hline
Real & 137 & 186 & 160 & 139 & 112 \\
Fake & 143 & 519 & 122 & 91 & 164 \\
Total & 280 & 705 & 282 & 230 & 276 \\
\hline
\end{tabular}
\\ 
\begin{tabular}{>{\centering}p{1.05cm}>{\centering}p{1.05cm}>{\centering}p{1.05cm}>{\centering}p{1.05cm}>{\centering}p{1.05cm}>{\centering\arraybackslash}p{1.05cm}}
\hline
Domain & Int. & Edu. & Soc. & Ent. & All \\
\hline
Real & 45 & 273 & 1,443 & 1,120 & 3,615 \\
Fake & 41 & 148 & 2,372 & 508 & 4,108 \\
Total & 86 & 421 & 3,815 & 1,628 & 7,723 \\
\hline
\end{tabular}
\label{tab:weibo_stats}
\end{table}

\textbf{Weibo21}~\cite{nan2021mdfend} is another Chinese multi-domain benchmark collected from Sina Weibo. It covers finance, health, military, science, politics, disaster, education, society, and entertainment. Its long-tailed domain distribution is reported in Table~\ref{tab:new_Weibo21_stats_full} and analyzed further in Appendix~\ref{appendix:domain_bias}.

\begin{table}[H]
\caption{Statistics of the original Weibo21 dataset. Dis., Edu., Soc., and Ent. denote disaster, education, society, and entertainment.}
\renewcommand{\arraystretch}{1.0}
\centering
\small
\setlength{\tabcolsep}{3pt}
\begin{tabular}{>{\centering}p{1.05cm}>{\centering}p{1.05cm}>{\centering}p{1.05cm}>{\centering}p{1.05cm}>{\centering}p{1.05cm}>{\centering\arraybackslash}p{1.05cm}}
\hline
Domain & Finance & Health & Military & Science & Politics \\
\hline
Real & 959 & 485 & 117 & 143 & 304 \\
Fake & 362 & 515 & 222 & 93 & 546 \\
All & 1,321 & 1,000 & 339 & 236 & 850 \\
\hline
\end{tabular}
\begin{tabular}{>{\centering}p{1.05cm}>{\centering}p{1.05cm}>{\centering}p{1.05cm}>{\centering}p{1.05cm}>{\centering}p{1.05cm}>{\centering\arraybackslash}p{1.05cm}}
\hline
Domain & Dis. & Edu. & Soc. & Ent. & All\\
\hline
Real & 184 & 243 & 1,198 & 1,007 & 4,640 \\
Fake & 591 & 248 & 1,417 & 440 & 4,434 \\
All & 775 & 491 & 2,615 & 1,447 & 9,074 \\
\hline
\end{tabular}
\label{tab:new_Weibo21_stats_full}
\end{table}

\section{Baselines}
\label{sec:baseline}
We compare \textsc{EGMD} with eight representative multimodal fake news detectors spanning general multimodal fusion, cross-modal alignment, frequency-domain modeling, domain adaptation, and MoE architectures.

\subsection{General Multimodal Detection}

\textbf{MFAN}~\cite{zheng2022mfan} proposed a multimodal graph structure that fuses text, images, and social graphs to boost detection performance.

\textbf{MRML}~\cite{peng2023mrml} investigated intra-modal and cross-modal relationships through triplet and contrastive learning techniques.

\textbf{COOLANT}~\cite{wang2023cross} enhanced text-image alignment by applying cross-modal contrastive learning alongside consistency-driven tasks.

\subsection{Cross-Domain and Expert-Based Detection}

\textbf{BMR}~\cite{ying2023bootstrapping} modeled news features from multiple views through bootstrap multi-view representations and utilized the Mixture of Experts network for multi-view fusion.

\textbf{FSRU}~\cite{lao2024frequency} converted spatial features into frequency spectrum features using Fourier transforms, extracting effective information through single-modality spectrum compression and cross-modality spectrum collaborative selection modules.

\textbf{MMDFND}~\cite{tong2024mmdfnd} employed a hierarchical extraction network with fused domain embeddings and attention mechanisms to achieve domain-adaptive modeling, fully integrating information from different modalities and domains via pivot transformations and adaptive normalization.

\textbf{MiMOE-FND}~\cite{liu2025modality} extended the MoE architecture by incorporating an interaction gating mechanism, which allowed the model to capture complex interactions and dependencies between modalities.

\textbf{DAMMFND}~\cite{lu2025dammfnd} addressed negative transfer in cross-domain settings via domain information decoupling. By incorporating domain-aware multi-view discrimination and a domain-augmented weighted decision strategy, it adaptively captured the heterogeneous contributions of multimodal features across domains, leading to more accurate and robust cross-domain fake news detection.

\subsection{Scope of Concurrent 2026 Methods}

We also examine two concurrent methods but do not copy their published numbers into our tables because the required supervision or reported outputs are not directly comparable. ADOSE~\cite{chen2026adose} is an active adaptation method that labels 10\% of the target-domain samples and evaluates leave-one-domain-out transfer over four selected Weibo domains; this additional supervision is unavailable to all methods in our setting. MemiMoE-FND~\cite{meng2026memimoe} evaluates standard multimodal detection on Weibo, Pheme, and PolitiFact without reporting domain-wise predictions for disparity evaluation. We therefore cite both methods as closely related work while restricting the quantitative comparison to methods rerun under identical splits, target-label access, and metrics.

\section{Evaluation Metrics}
\label{sec:evaluation metrics}
Following prior work~\cite{liu2021authors, li2024dual}, we adopt accuracy to evaluate overall performance, False Positive Equality Difference (FPED), and False Negative Equality Difference (FNED)~\cite{dixon2018measuring, zhao2024universal} to assess fairness.

\textbf{Predictive performance.}
Accuracy is the fraction of correctly classified samples over the complete evaluation set.

\textbf{Cross-domain disparity.}
FPED sums the absolute differences between each domain-specific false-positive rate and the global false-positive rate. FNED is defined analogously using false-negative rates:

\begin{equation}
FPED = \sum_{d \in K} \left| FPR - FPR_d \right|
\label{eq:FPED}
\end{equation}
\begin{equation}
FNED = \sum_{d \in K} \left| FNR - FNR_d \right|
\label{eq:FNED}
\end{equation}
where $K$ denotes the set of domains. We define $\text{Total}=\text{FPED}+\text{FNED}$, with lower values indicating smaller cross-domain error disparities.

\textbf{Joint performance-fairness score.}
PFS provides a secondary scalar summary of Accuracy and Total:
\begin{equation}
PFS = \theta \cdot \text{ACC} + (1-\theta) \cdot e^{-2 \cdot \text{Total}},
\end{equation}
where $\theta=0.8$ encodes an accuracy-prioritized reporting preference. Its partial derivatives,
$\partial\mathrm{PFS}/\partial\mathrm{ACC}=\theta>0$ and
$\partial\mathrm{PFS}/\partial\mathrm{Total}=-2(1-\theta)e^{-2\mathrm{Total}}<0$,
make the intended directions explicit. The factor 2 fixes the curvature of the bounded disparity term. Neither constant is claimed to be theoretically optimal, and a scalar score may obscure cases in which Accuracy and disparity move in opposite directions. Accordingly, all tables report Accuracy, FPED, FNED, and Total separately; PFS is used only as a supplementary summary rather than as standalone evidence of superiority.

\section{Extended Related Work}
\label{app:extended_related_work}

This section provides a detailed review of traditional unimodal approaches and the evolution of multimodal fake news detection architectures.

\subsection{Traditional Fake News Detection}
Traditional methods primarily rely on unimodal information such as text or propagation structures.
\begin{itemize}
    \item \textbf{Generalization Strategies:} ENDEF~\cite{zhu2022generalizing} applied causal inference to mitigate entity bias. REAL-FND~\cite{mosallanezhad2022domain} introduced reinforcement learning to adjust feature representations across domains based on user-news interactions.
    \item \textbf{Complex Modeling:} UPFD~\cite{dou2021user} incorporated user posting histories, while M³FEND~\cite{9802916} addressed label scarcity through multi-view modeling and memory mechanisms.
    \item \textbf{Debiasing Frameworks:} Notably, DTDBD~\cite{li2024dual} provided a dual-teacher distillation framework to enhance cross-domain performance. However, as these methods are designed for unimodal inputs, they lack the flexibility to generalize to complex multimodal scenarios where visual context is crucial.
\end{itemize}

\subsection{Multimodal Fake News Detection}
Multimodal detection integrates text and images to improve accuracy, with research focusing on intra- and cross-modal dynamics.
\begin{itemize}
    \item \textbf{Fusion Mechanisms:} MFAN~\cite{zheng2022mfan} proposed a graph-based fusion of text, images, and social networks. MRML~\cite{peng2023mrml} and COOLANT~\cite{wang2023cross} utilized triplet and contrastive learning to enhance alignment. More recently, FSRU~\cite{lao2024frequency} introduced spectrum-based fusion using Fourier transforms, and MTS~\cite{sun2025multimodal} leveraged Taylor series expansion for scalable interaction modeling.
    \item \textbf{Domain Adaptation:} MMDFND~\cite{tong2024mmdfnd} employs fused domain embeddings and attention mechanisms for domain-adaptive modeling, whereas ADOSE~\cite{chen2026adose} actively acquires labels from a target domain for multi-source adaptation.
    \item \textbf{Mixture-of-Experts (MoE):} To handle heterogeneity, BMR~\cite{ying2023bootstrapping} uses MoE for reweighted fusion, MiMOE-FND~\cite{liu2025modality} models modality interactions, and MemiMoE-FND~\cite{meng2026memimoe} introduces learnable memory slots and hierarchical expert fusion.
\end{itemize}
Despite these advances, existing multimodal MoE methods typically focus on fusion efficacy. In contrast, \textsc{EGMD} uses expert guidance to reduce domain-specific bias while preserving shared semantic cues.

\section{Domain Bias Analysis in Weibo21}
\label{appendix:domain_bias}

\begin{figure}[H]
    \centering
    \includegraphics[width=\columnwidth]{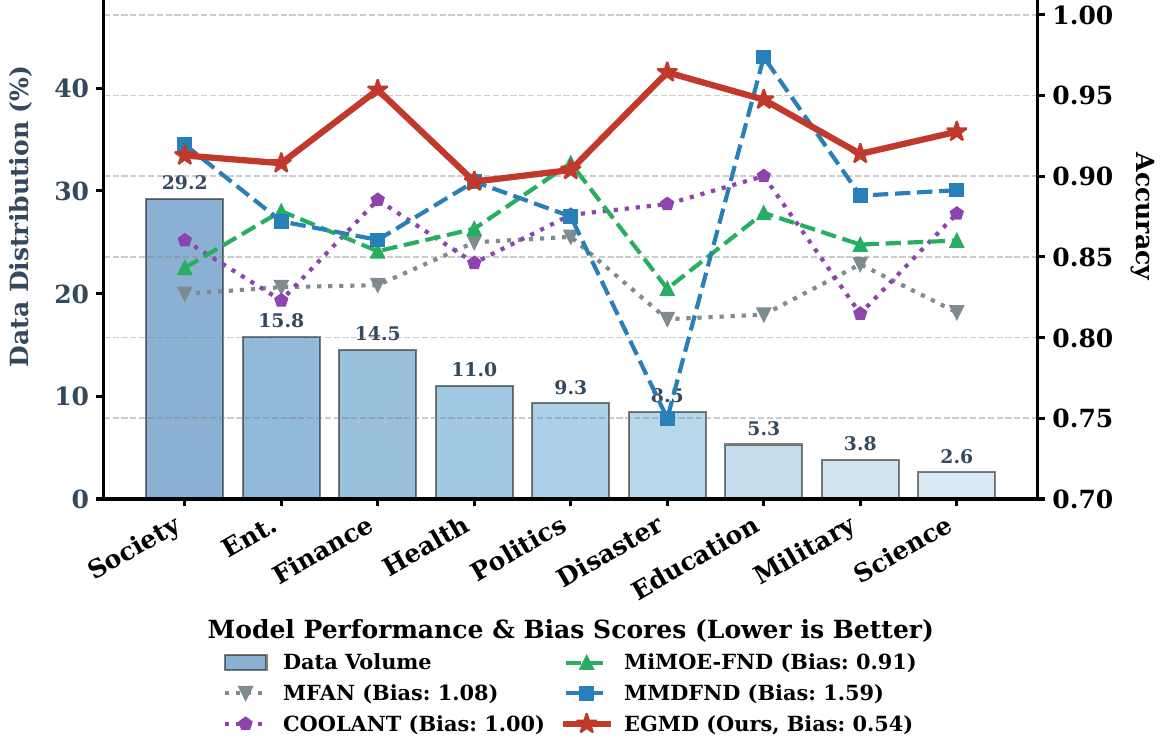}
    \caption{Performance comparison between head and tail domains on Weibo21, illustrating cross-domain disparity in multimodal fake news detection.}
\label{fig:motivation_analysis}
\end{figure}

This section provides a statistical examination of the domain imbalances in the Weibo21 dataset~\cite{nan2021mdfend}, as visualized in Figure~\ref{fig:motivation_analysis}.

\paragraph{Data Volume Discrepancy.}
The dataset exhibits an $11.08\times$ disparity in data volume between the most represented domain (\textit{Society}, $28.8\%$) and the least represented domain (\textit{Science}, $2.6\%$). The top three domains (Society, Entertainment, and Finance) collectively account for 59.3\% of the data, whereas the bottom three account for 11.7\%. This long-tailed distribution can bias optimization toward majority domains and confounds domain shift with unequal sample size.

\paragraph{Distributional Independence.}
Crucially, the prevalence of fake news varies significantly across domains (ranging from $27.4\%$ in \textit{Finance} to $76.1\%$ in \textit{Disaster}) but shows a weak negative correlation ($r=-0.179$) with data volume. This finding indicates that the volume imbalance is not a sampling artifact but a structural property. Consequently, simple volumetric heuristics fail; effective detection requires learning robust, domain-invariant semantics rather than relying on domain-specific statistics.

\paragraph{Implications for Fairness.}
The severe imbalance makes global accuracy insufficient because errors in small domains contribute little to the aggregate score. On Weibo21, \textsc{EGMD} achieves a Total bias of 0.5415, compared with 1.5860 for MMDFND, while maintaining the strongest overall accuracy. This result motivates reporting domain-level performance and disparity metrics together.

\section{\textsc{Weibo\_Balanced}: Construction and Full Results}
\label{app:dataset_details}

In this section, we provide a comprehensive description of the construction, statistics, and validation of the \textsc{Weibo\_Balanced} dataset.

\subsection{Motivation: Controlling Data-Volume Imbalance}
The original Weibo21 dataset exhibits severe domain imbalance (Table~\ref{tab:new_Weibo21_stats_full}). Consequently, results on the original benchmark reflect both domain shift and unequal data volume. We construct \textsc{Weibo\_Balanced} to reduce this confounding effect and provide a more controlled evaluation of cross-domain performance.

\subsection{Data Collection and Quality Assurance}
\textbf{Collection strategy.}
We follow the collection pipeline of Weibo21~\cite{nan2021mdfend}. Seed keywords are derived from high-frequency terms in underrepresented domains and used for targeted crawling. For the dominant \textit{Society} domain, we apply random downsampling to retain a representative subset while reducing its prevalence.

\textbf{Annotation protocol.}
Ten domain experts independently annotate each newly collected instance. A domain label is accepted when at least eight experts agree. Instances below this threshold undergo further review until consensus is reached.

\subsection{Dataset Statistics}
Table~\ref{tab:new_Weibo_Balanced_stats_full} reports the resulting distribution. Relative to Weibo21, the balanced benchmark adds 5,314 instances and removes 988, yielding a net increase of 4,326 instances. Domain totals range from 1,282 to 1,627, reducing the maximum ratio to $1.27\times$.

\begin{table}[H]
\caption{Statistics of the constructed \textsc{Weibo\_Balanced} dataset. Dis., Edu., Soc., and Ent. denote disaster, education, society, and entertainment.}
\renewcommand{\arraystretch}{1.0}
\centering
\small
\setlength{\tabcolsep}{3pt}
\begin{tabular}{>{\centering}p{1.05cm}>{\centering}p{1.05cm}>{\centering}p{1.05cm}>{\centering}p{1.05cm}>{\centering}p{1.05cm}>{\centering\arraybackslash}p{1.05cm}}
\hline
Domain & Finance & Health & Military & Science & Politics \\
\hline
Real & 1,033 & 662 & 1,057 & 1,023 & 792 \\
Fake & 430 & 862 & 325 & 259 & 822 \\
All & 1,463 & 1,524 & 1,382 & 1,282 & 1,614 \\
\hline
\end{tabular}
\\
\begin{tabular}{>{\centering}p{1.05cm}>{\centering}p{1.05cm}>{\centering}p{1.05cm}>{\centering}p{1.05cm}>{\centering}p{1.05cm}>{\centering\arraybackslash}p{1.05cm}}
\hline
Domain & Dis. & Edu. & Soc. & Ent. & All\\
\hline
Real & 721 & 928 & 714 & 1,007 & 7,937 \\
Fake & 867 & 545 & 913 & 440 & 5,463 \\
All & 1,588 & 1,473 & 1,627 & 1,447 & 13,400 \\
\hline
\end{tabular}
\label{tab:new_Weibo_Balanced_stats_full}
\end{table}

\subsection{Per-Domain Results}
Table~\ref{tab:performance comparison with Weibo Balanced Dataset} reports the per-domain results underlying the \textsc{Weibo\_Balanced} comparison summarized in the main paper.

\begin{table}[H]
\caption{Results on \textsc{Weibo\_Balanced}. Best and second-best values are bold and underlined.}
\label{tab:performance comparison with Weibo Balanced Dataset}
\centering
\scriptsize
\renewcommand{\arraystretch}{1.05}
\setlength{\tabcolsep}{3.2pt}

\begin{tabular}{lccccc}
\toprule
Method & Fin. & Health & Mil. & Sci. & Pol. \\
\midrule
FSRU & 0.7737 & 0.7367 & 0.8064 & 0.8337 & 0.8549 \\
MMDFND & 0.8980 & \textbf{0.9340} & \underline{0.9286} & 0.9110 & \underline{0.9261} \\
MiMOE-FND & 0.9075 & 0.8852 & 0.9152 & \underline{0.9375} & 0.9081 \\
DAMMFND & \textbf{0.9363} & 0.8500 & 0.9184 & 0.8743 & 0.9224 \\
\rowcolor{black!10}\textbf{\textsc{EGMD}} & \underline{0.9235} & \underline{0.9057} & \textbf{0.9326} & \textbf{0.9581} & \textbf{0.9304} \\
\bottomrule
\end{tabular}

\vspace{3pt}
\begin{tabular}{lcccc}
\toprule
Method & Dis. & Edu. & Soc. & Ent. \\
\midrule
FSRU & 0.7915 & 0.7520 & 0.8111 & 0.8444 \\
MMDFND & 0.9189 & \underline{0.9273} & 0.8974 & 0.8989 \\
MiMOE-FND & 0.9152 & 0.9116 & 0.8877 & 0.9258 \\
DAMMFND & \textbf{0.9221} & 0.8973 & \textbf{0.9292} & \textbf{0.9667} \\
\rowcolor{black!10}\textbf{\textsc{EGMD}} & \underline{0.9148} & \textbf{0.9445} & \underline{0.9103} & \underline{0.9366} \\
\bottomrule
\end{tabular}

\vspace{3pt}
\begin{tabular}{lccccc}
\toprule
Method & Acc. $\uparrow$ & FNED $\downarrow$ & FPED $\downarrow$ & Total $\downarrow$ & PFS $\uparrow$ \\
\midrule
FSRU & 0.8002 & \textbf{0.1366} & 0.6531 & \underline{0.7897} & 0.6814 \\
MMDFND & \underline{0.9156} & 0.8512 & 0.2999 & 1.1511 & 0.7525 \\
MiMOE-FND & 0.9070 & 0.6041 & \underline{0.1931} & 0.7933 & 0.7256 \\
DAMMFND & 0.9134 & 0.6678 & 0.2615 & 0.9293 & \underline{0.7619} \\
\rowcolor{black!10}\textbf{\textsc{EGMD}} & \textbf{0.9280\textsuperscript{\dag}} & \underline{0.1491} & \textbf{0.1885} & \textbf{0.3376\textsuperscript{\dag}} & \textbf{0.8442\textsuperscript{\dag}} \\
\bottomrule
\end{tabular}
\end{table}

\textbf{Results.}
\textsc{EGMD} obtains the highest overall Accuracy (0.9280) and PFS (0.8442), improving over the strongest competing values by 0.0124 and 0.0823, respectively. It also reduces Total bias to 0.3376, which is 0.4521 below the next-lowest value. These results confirm that the gains summarized in the main paper persist after substantially reducing domain-size imbalance.

\section{Additional Ablation, Robustness, and Efficiency Results}
\label{app:ablation_robustness}

This section supplements the main ablation with progressive component construction, deployment without ground-truth domain labels, and computational efficiency.

\subsection{Progressive Component Construction}

Starting from the coherence-aware gating mechanism alone gives an Accuracy of 0.8906, a \textit{Total} score of 0.9332, and a PFS of 0.7434. Adding DDN yields the largest early reduction in \textit{Total}, lowering it to 0.7725. The MoE teacher and mutual distillation then reduce \textit{Total} to 0.6983 and 0.5938, respectively, before the complete model reaches 0.5415. This bottom-up trend complements the leave-one-component-out ablation in the main paper.

\subsection{Robustness to Missing Domain Labels}
\label{sec:label_free}

\textbf{Evaluation protocol.}
When domain labels are unavailable at inference, \textsc{EGMD} assigns each instance to a domain using the learned prototypes described in the main paper's joint-optimization section. We compare this practical setting with two references: an oracle using ground-truth domain labels and a lower bound using random domain assignment. This comparison isolates the effect of routing quality without changing the trained model.

\begin{table}[H]
\caption{Robustness to missing domain labels on Weibo21 under oracle, prototype-inferred, and random routing.}
\centering
\small
\setlength{\tabcolsep}{4pt}
\begin{tabular}{lccc}
\toprule
Setting & Domain source & Accuracy $\uparrow$ & Total $\downarrow$\\
\midrule
Oracle & Ground-truth & 0.9163 & 0.5415\\
\textsc{EGMD} (Ours) & Prototype-inferred & 0.9063 & 0.6137\\
Lower bound & Random & 0.8927 & 0.7230\\
\bottomrule
\end{tabular}
\label{tab:label_free}
\end{table}

\textbf{Results.}
As shown in Table~\ref{tab:label_free}, prototype-inferred routing reaches 0.9063 Accuracy and 0.6137 Total. Relative to the oracle, it decreases Accuracy by 0.0100 and increases Total by 0.0722. Nevertheless, it remains substantially stronger than random routing, improving Accuracy by 0.0136 and reducing Total by 0.1093. These results show that accurate domain routing remains beneficial, while learned prototypes provide a reliable fallback when explicit domain labels are unavailable. This fallback performs closed-set routing among the $K$ source-domain branches; it does not create a branch for a novel or redefined domain taxonomy.

\subsection{Computational Efficiency}

The teacher, prototype updates, mutual learning, and both distillation losses supervise training, whereas inference activates only one student branch. Table~\ref{tab:structure_efficiency} therefore reports both resource cost and Weibo21 effectiveness. Relative to MMDFND, \textsc{EGMD} uses 17.3M more parameters (15.9\%), adds 11.2 seconds per epoch (24.8\%), and increases single-instance latency by 3.1 milliseconds (18.8\%). This increase accompanies a 0.0832 gain in PFS and a 1.0445 reduction in Total bias.

\begin{table}[H]
\caption{Efficiency and effectiveness on Weibo21. Training time is measured per epoch on an NVIDIA RTX 4090; latency is measured per instance.}
\centering
\scriptsize
\setlength{\tabcolsep}{2.5pt}
\begin{tabular}{lccccc}
\toprule
Model & Params & Train & Lat. & Total $\downarrow$ & PFS $\uparrow$\\
\midrule
MMDFND & \textbf{108.5M} & \textbf{45.2s} & \textbf{16.5ms} & 1.5860 & 0.7176\\
MiMOE-FND & 112.4M & 51.8s & 18.2ms & 0.9058 & 0.7263\\
\rowcolor{black!10}\textbf{\textsc{EGMD}} & 125.8M & 56.4s & 19.6ms & \textbf{0.5415} & \textbf{0.8008}\\
\bottomrule
\end{tabular}
\label{tab:structure_efficiency}
\end{table}

\section{Cross-Dataset Generalization: Per-Domain Results}
\label{app:per_domain_45}

This section provides the full per-domain breakdown underlying the cross-dataset comparison summarized in the main paper.
\begin{table}[H]
\caption{Per-domain and overall results when training on Weibo and testing on Weibo21.}
\label{tab:cross_datasets_combined}
\centering
\scriptsize
\renewcommand{\arraystretch}{1.05}
\setlength{\tabcolsep}{3.2pt}

\begin{tabular}{lccccc}
\toprule
Method & Fin. & Health & Mil. & Sci. & Pol. \\
\midrule
FSRU & 0.6211 & 0.6858 & 0.6513 & 0.6333 & 0.6047 \\
MMDFND & 0.6979 & \underline{0.7784} & 0.7071 & 0.6835 & 0.7746 \\
DAMMFND & \underline{0.7092} & \textbf{0.7871} & \underline{0.7626} & \underline{0.7194} & \textbf{0.8144} \\
\rowcolor{black!10}\textbf{\textsc{EGMD}} & \textbf{0.7952} & 0.7766 & \textbf{0.7980} & \textbf{0.7482} & \underline{0.7955} \\
\bottomrule
\end{tabular}

\vspace{3pt}
\begin{tabular}{lcccc}
\toprule
Method & Dis. & Edu. & Soc. & Ent. \\
\midrule
FSRU & 0.5835 & 0.6250 & 0.6779 & 0.6414 \\
MMDFND & 0.6909 & 0.7208 & 0.7726 & \underline{0.7967} \\
DAMMFND & \textbf{0.7863} & \underline{0.7633} & \underline{0.7848} & 0.7523 \\
\rowcolor{black!10}\textbf{\textsc{EGMD}} & \underline{0.7801} & \textbf{0.7986} & \textbf{0.7945} & \textbf{0.8143} \\
\bottomrule
\end{tabular}

\vspace{3pt}
\begin{tabular}{lccccc}
\toprule
Method & Acc. $\uparrow$ & FNED $\downarrow$ & FPED $\downarrow$ & Total $\downarrow$ & PFS $\uparrow$ \\
\midrule
FSRU & 0.6447 & 1.1327 & 0.6890 & 1.8217 & 0.5210 \\
MMDFND & 0.7528 & 0.6151 & 0.5886 & 1.2037 & 0.6202 \\
DAMMFND & \underline{0.7674} & \underline{0.3343} & \underline{0.2941} & \underline{0.6284} & \underline{0.6708} \\
\rowcolor{black!10}\textbf{\textsc{EGMD}} & \textbf{0.7944\textsuperscript{\dag}} & \textbf{0.2265\textsuperscript{\dag}} & \textbf{0.2363\textsuperscript{\dag}} & \textbf{0.4628\textsuperscript{\dag}} & \textbf{0.7148\textsuperscript{\dag}} \\
\bottomrule
\end{tabular}
\end{table}

\FloatBarrier

\textbf{Results.}
\textsc{EGMD} achieves the highest accuracy in six of the nine target domains and improves overall Accuracy from 0.7674 to 0.7944 relative to DAMMFND. It also reduces Total bias from 0.6284 to 0.4628 and raises PFS from 0.6708 to 0.7148. The simultaneous improvement in domain-level accuracy and aggregate disparity supports the cross-dataset conclusion in the main paper.

\section{Feature Visualization}
\label{app:feature_visualization}

\textbf{Purpose and setup.}
We use t-SNE~\cite{van2008visualizing} to compare raw multimodal features with representations learned by MMDFND and \textsc{EGMD} on Weibo. The top row is colored by veracity label to assess class separation, and the bottom row is colored by domain to inspect domain-dependent structure. Because t-SNE is projection-dependent, this analysis is treated as qualitative supporting evidence.

\begin{figure}[H]
\centering
\begin{subfigure}{0.14\textwidth}
    \centering
    \includegraphics[width=\linewidth]{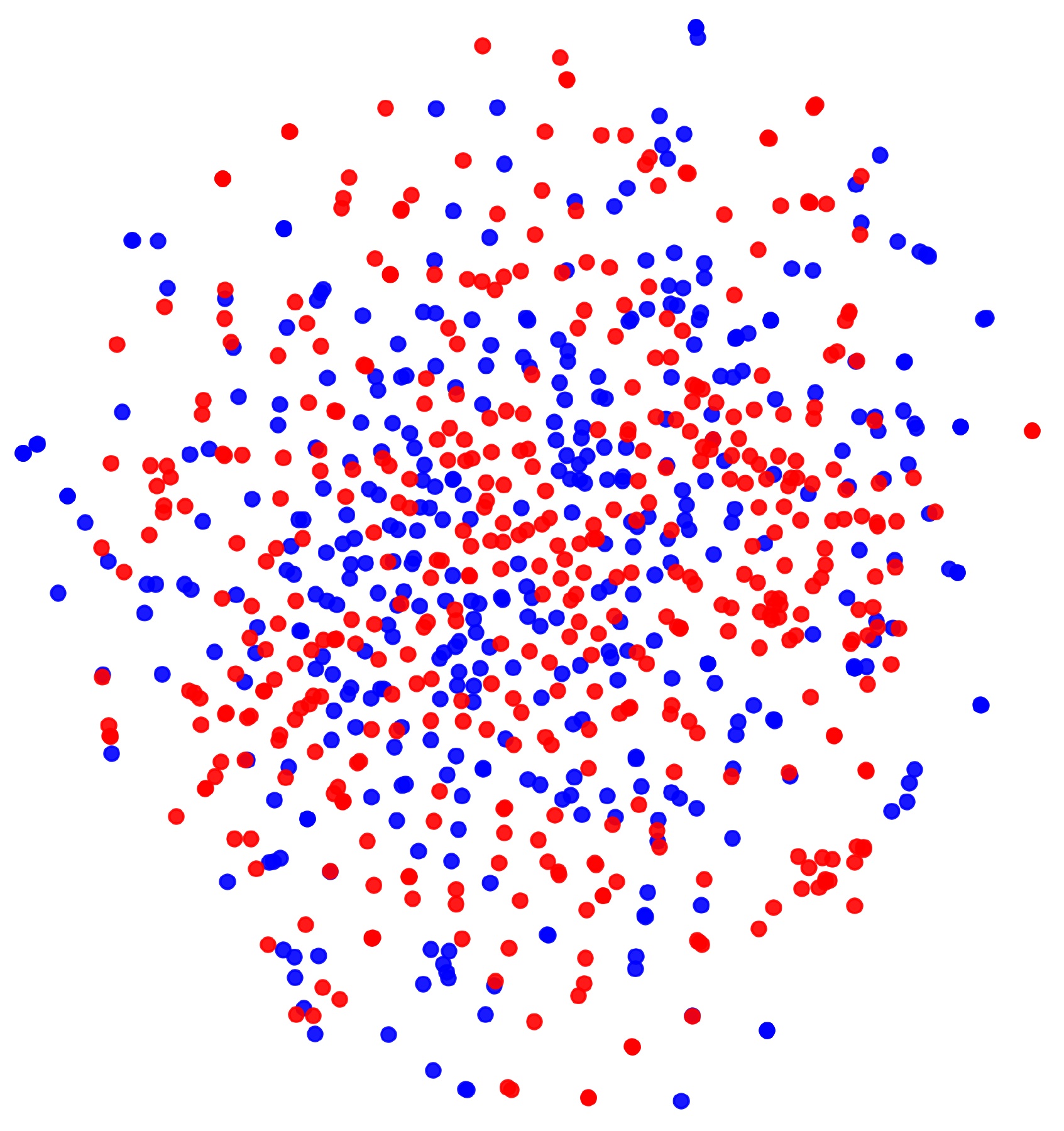}
    \caption{Original}
\end{subfigure}
\hfill
\begin{subfigure}{0.14\textwidth}
    \centering
    \includegraphics[width=\linewidth]{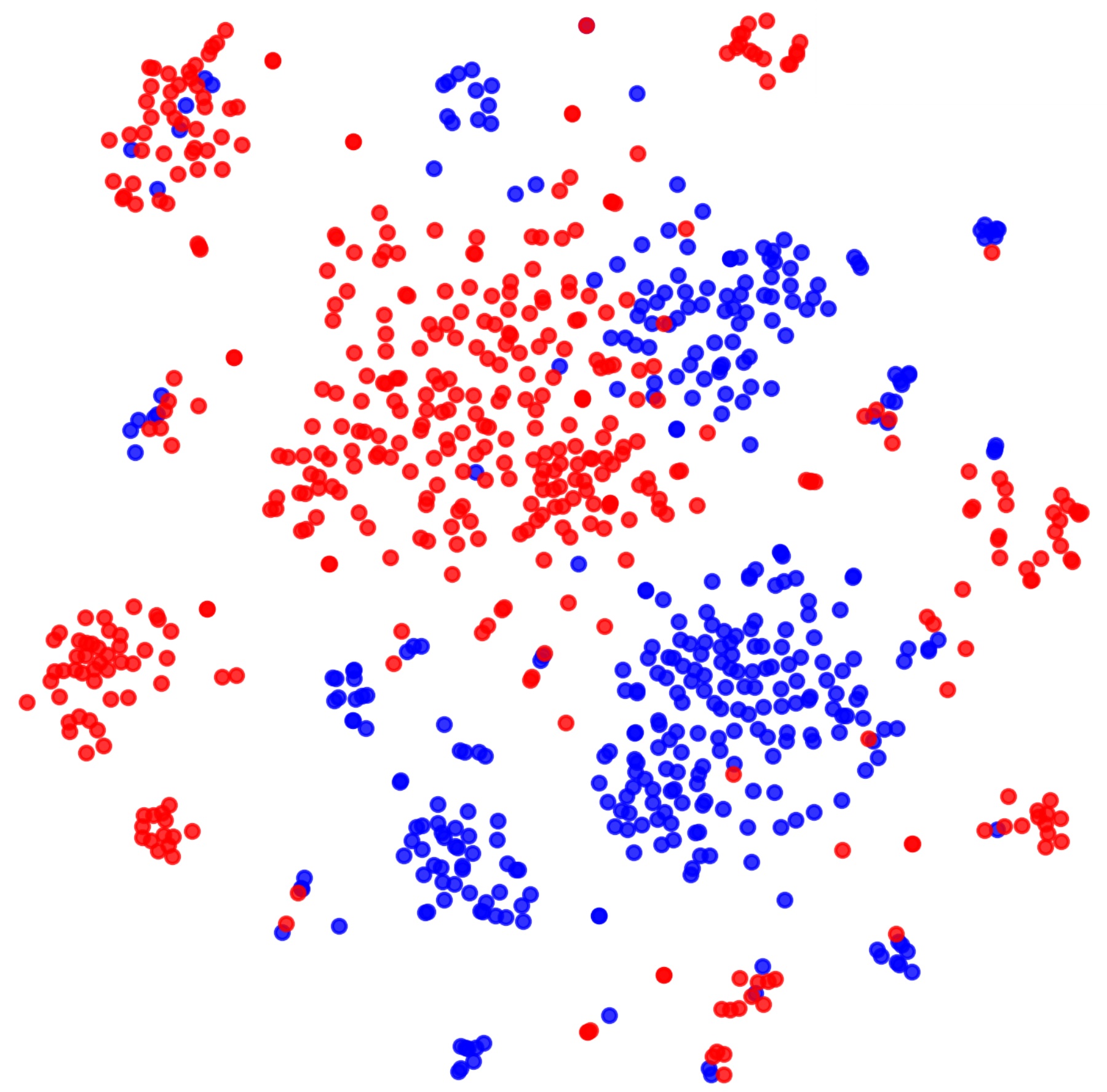}
    \caption{MMDFND}
\end{subfigure}
\hfill
\begin{subfigure}{0.14\textwidth}
    \centering
    \includegraphics[width=\linewidth]{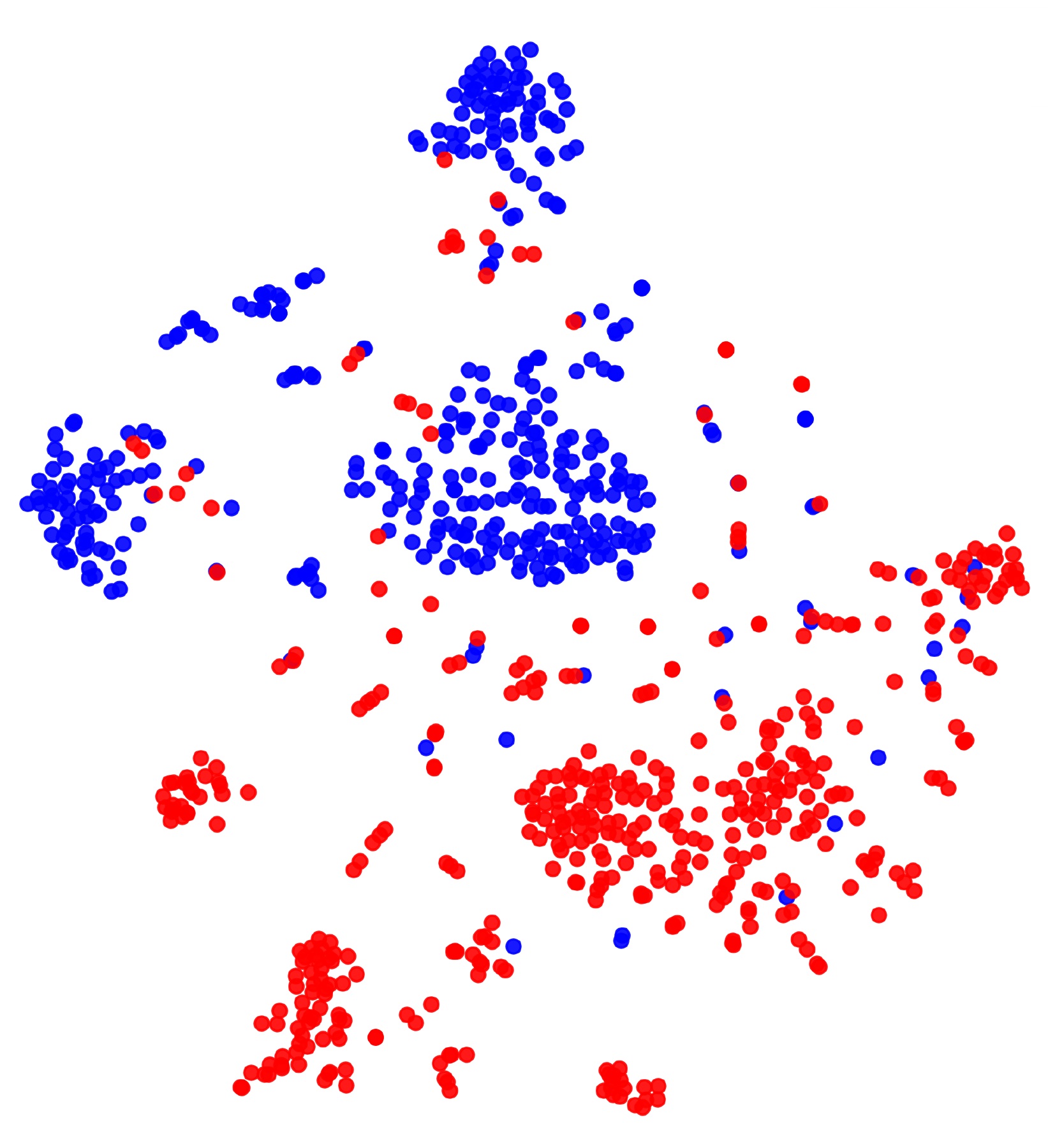}
    \caption{\textsc{EGMD}}
\end{subfigure}

\medskip

\begin{subfigure}{0.14\textwidth}
    \centering
    \includegraphics[width=\linewidth]{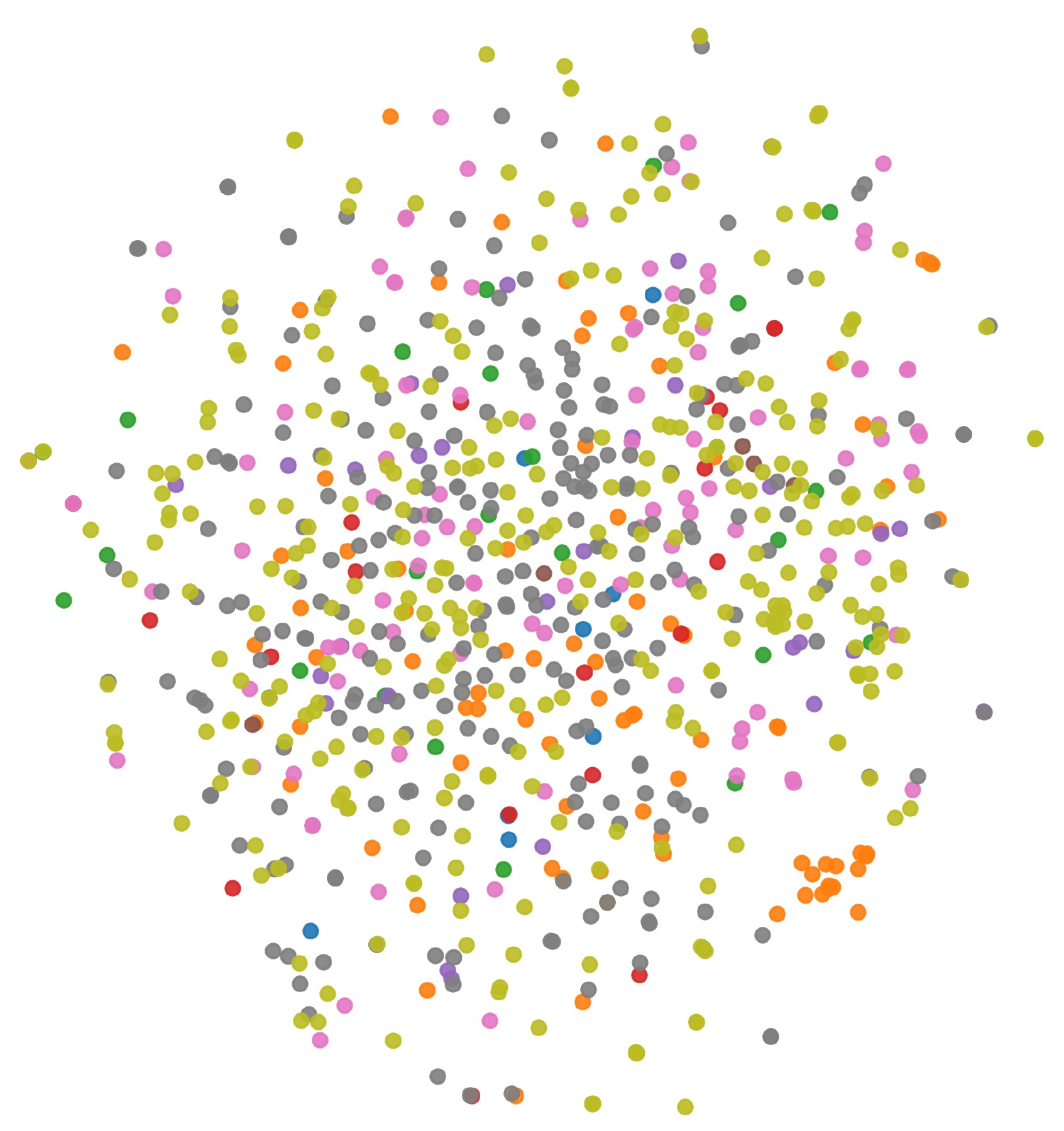}
    \caption{Original}
\end{subfigure}
\hfill
\begin{subfigure}{0.14\textwidth}
    \centering
    \includegraphics[width=\linewidth]{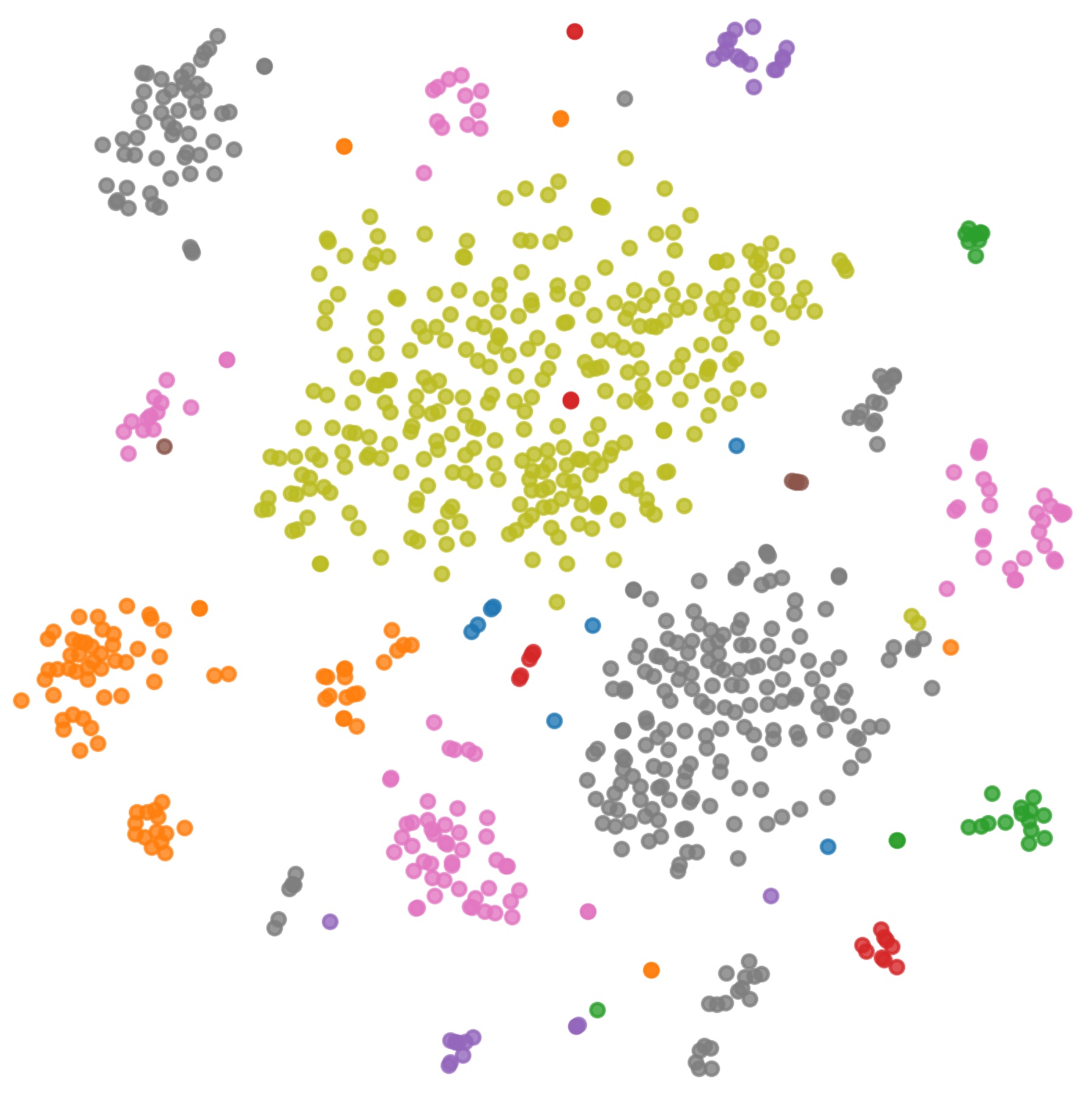}
    \caption{MMDFND}
\end{subfigure}
\hfill
\begin{subfigure}{0.14\textwidth}
    \centering
    \includegraphics[width=\linewidth]{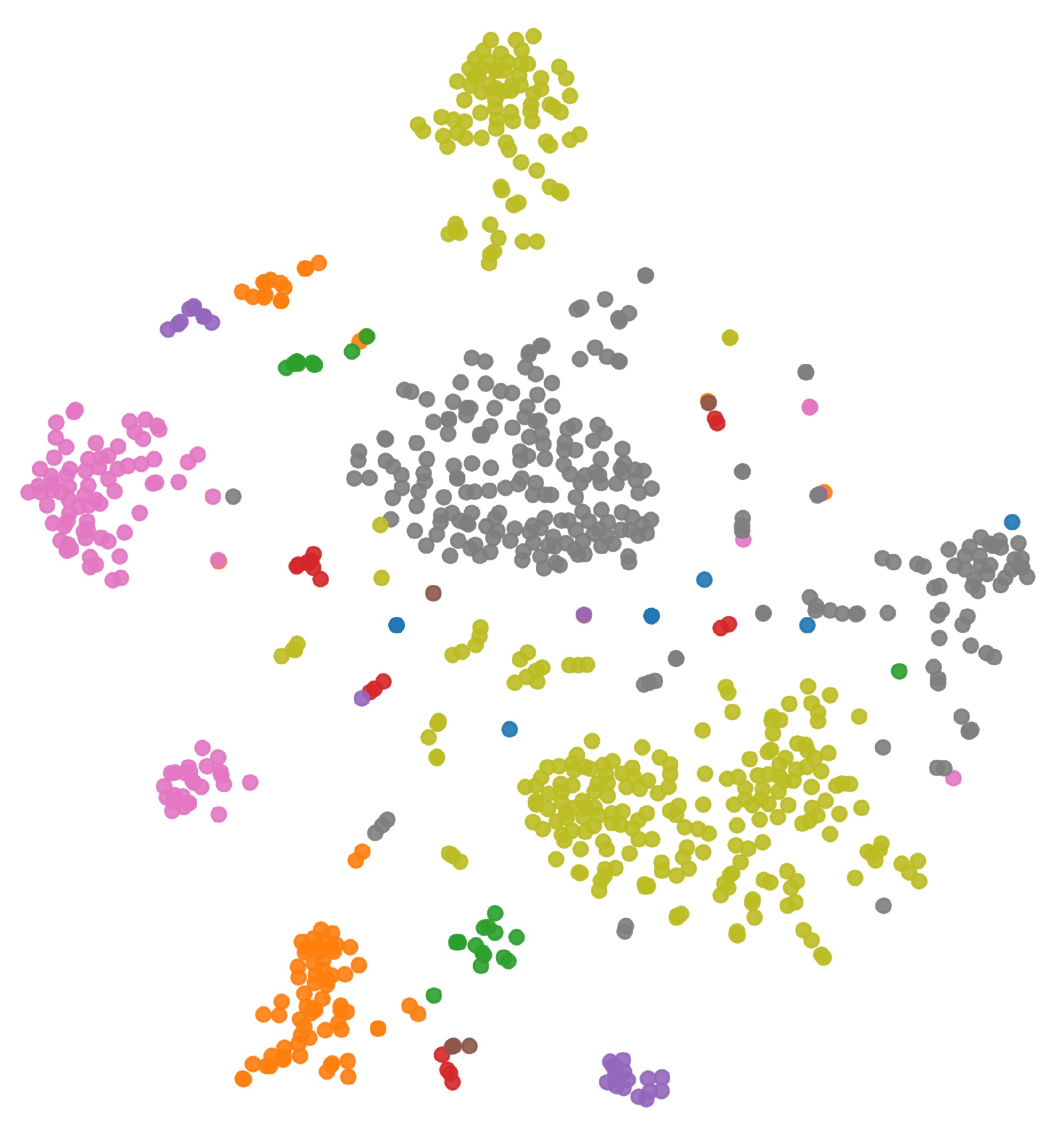}
    \caption{\textsc{EGMD}}
\end{subfigure}
\caption{t-SNE visualization of feature distributions on Weibo. The top row is colored by veracity label and the bottom row by news domain.}
\label{fig:tsne}
\end{figure}

\textbf{Results.}
As shown in Figure~\ref{fig:tsne}, raw features exhibit substantial overlap under both colorings. MMDFND improves local grouping but retains mixed regions. \textsc{EGMD} produces more compact veracity clusters while reducing the visually dominant domain-specific grouping, consistent with the quantitative disparity results.

\FloatBarrier

\end{document}